\documentclass[article,journal]{IEEEtran}
\usepackage{amsmath,amssymb,amsfonts}
\usepackage[linesnumbered,ruled]{algorithm2e}
\usepackage{listings}%
\usepackage{float}%
\usepackage{orcidlink}
\usepackage{multirow}%
\usepackage{amsthm}%
\usepackage{mathrsfs}%
\usepackage{xcolor}%
\usepackage{manyfoot}%
\usepackage{booktabs}%
\usepackage{caption}
\usepackage{bbm}
\usepackage{mathtools}
\usepackage{threeparttable}

\usepackage{array}
\usepackage[caption=false,font=footnotesize,labelfont=rm,textfont=rm]{subfig}\usepackage{textcomp}
\usepackage{stfloats}
\usepackage{url}
\usepackage{verbatim}
\usepackage{graphicx}
\def\BibTeX{{\rm B\kern-.05em{\sc i\kern-.025em b}\kern-.08em
    T\kern-.1667em\lower.7ex\hbox{E}\kern-.125emX}}
\usepackage{balance}
\usepackage{nomencl}

\makenomenclature

\newtheorem{myDef}{Definition}

\begin{document}
\title{Distributed Relative Localization Based on Ultra-WideBand and LiDAR for Multi-robot with Limited Communication}

\author{~~~Zhiqiang Cao\textsuperscript{\orcidlink{0000-0002-0823-7155}}, 
Ran Liu\textsuperscript{\orcidlink{0000-0002-6343-4645}}, \IEEEmembership{Senior Member, IEEE}, 
Billy Pik Lik Lau\textsuperscript{\orcidlink{0000-0001-5133-2791}}, \IEEEmembership{Senior Member, IEEE}, \\
Chau Yuen\textsuperscript{\orcidlink{0000-0002-9307-2120}}, \IEEEmembership{Fellow, IEEE},
and U-Xuan Tan\textsuperscript{\orcidlink{0000-0002-5757-1379}}, \IEEEmembership{Senior Member, IEEE}

\thanks{Z. Cao, B. P. L. Lau, and U-X. Tan are with the Engineering Product Development, Singapore University of Technology and Design, Singapore 487372. zhiqiang\_cao@mymail.sutd.edu.sg, and \{billy\_lau, uxuan\_tan\}@sutd.edu.sg.

R. Liu and C. Yuen are with the School of Electrical and Electronic Engineering, Nanyang Technological University, Singapore 639798. \{ran.liu, chau.yuen\}@ntu.edu.sg.

Corresponding author: Chau Yuen}

}

\markboth{IEEE PREPRINT, ~2024}%
{How to Use the IEEEtran \LaTeX \ Templates}

\maketitle

\begin{abstract}
Relative localization is crucial for a multi-robot system to collaboratively perform tasks, such as exploration and formation.
However, this is highly challenging for homogeneous robots with similar appearance in GPS-denied and communication-limited environments.
In this paper, we propose a fully distributed relative position estimation approach for a team of robots based on onboard UWB and LiDAR sensors, in which LiDAR is utilized to obtain the position of anonymous objects in Line-of-Sight (LOS), and UWB is used for ranging between robots.
We construct two graphs, namely UWB connection graph and LiDAR connection graph, to represent the spatial relationship among objects (robots and obstacles) based on UWB and LiDAR measurements.
Identification and relative position estimation are formulated as a common subgraph matching problem.
A falsely-matched robot identification approach is designed to recognize the falsely-matched results caused by obstacle blockage in LiDAR field of view.
These robots are then localized by leveraging the well-matched robots and the UWB ranging measurements in the UWB connection graph.
We conducted experiments to evaluate the performance of our approach.
The results show that the proposed approach is capable of achieving satisfactory positioning accuracy for a team of robots in a distributed manner with only exchanging limited information.

\end{abstract}



\section{Introduction}
\IEEEPARstart{M}{ultiple} robot systems have received a lot of attention from the robotics community in the last decades \cite{Introduction-multi-robot-system}.
There is evidence that multi-robot systems play a vital role in rescue and high-risk scenarios due to their collaborative capabilities, enhancing efficiency, coverage, and adaptability compared to single-robot systems.
NASA released a Cooperative Autonomous Distributed Robotic Exploration (CADRE) project \cite{Introduction-CADRE} that aims to develop a network of small mobile robots that improve autonomous robot swarm exploration on the Moon, Mars, and beyond.

\begin{figure}
\centering
\includegraphics[width=90mm]{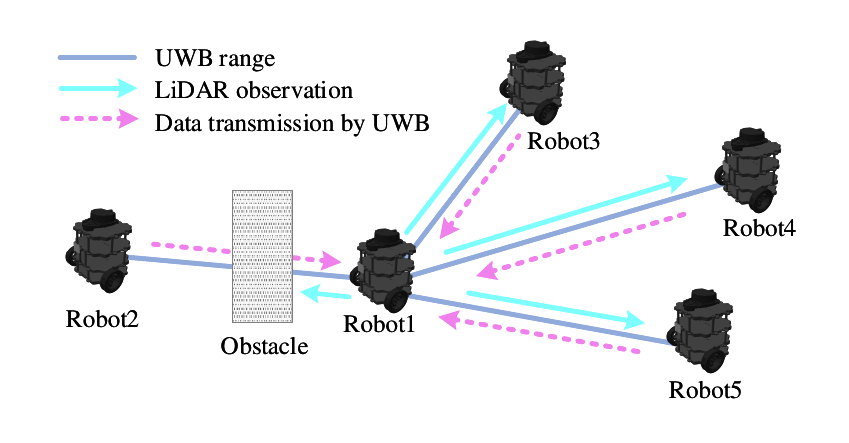}
\caption{Overview of the proposed distributed relative localization system using Robot1 as an example. Each robot is equipped with one UWB node to measure the distances to nearby robots and broadcast a small amount of UWB ranging data. The robot also carries LiDAR to estimate the position of the surrounding objects (robots and obstacles) in LOS. The proposed relative localization approach independently runs in a distributed manner.}
\label{figure-overview}
\end{figure}

The relative position between robots is an essential component for multi-robot system to execute decision-making \cite{Introduction-decision-making} and collaborative operations \cite{Introduction-cooperative-operation}.
In general, the system framework can be divided into two main categories, namely centralized and distributed \cite{Introduction-relative-category}.
The centralized approach is deployed on a server, requiring intensive communication and vulnerability to any server failure \cite{Introduction-centralized}.
In the distributed approach, each robot performs pose estimation based on local measurements and a small amount of data shared by neighboring robots.
Compared to the centralized system, the distributed system is well-suited for situations with limited communication bandwidth \cite{Introduction-distributed}.

At present, most of the relative localization solutions focus on Light Detection and Ranging (LiDAR) or visual cameras.
Both solutions have proven feasible in achieving accurate relative position estimation for objects in Line-of-Sight (LOS) scenarios \cite{Introduction-LiDAR-camera}.
However, the robots' similar appearance poses challenges to correctly identifying teammates in a multi-robot system \cite{Introduction-robot-id}.
Compared to LiDAR and visual cameras, Ultra-WideBand (UWB) can provide unique identification information for each node.
Meanwhile, it provides centimeter-level peer-to-peer ranging accuracy in ideal conditions and can also transmit a small amount of data to surrounding UWB nodes \cite{Introduction-UWB} \cite{Introduction-UWB-2}.
These features are vital for a multi-robot system with similar appearance working in communication-constrained scenes.
However, the need for bearing information among nodes affects UWB's practicality in localization.
Some researchers utilized the Angle of Arrival (AoA) technique to estimate the bearing \cite{AoA-1}\cite{AoA-2} for UWB localization.
Nevertheless, AoA solutions usually require rigorous clock synchronization.

This paper presents a distributed relative localization approach for a team of robots with similar appearance, employing UWB and LiDAR sensors without any external communication and positioning infrastructures, as shown in Fig. \ref{figure-overview}.
The UWB sensor is utilized to measure the range between robots and broadcast a few ranging data to nearby robots, while the LiDAR detects and localizes the surrounding objects in LOS.
We first construct UWB and LiDAR connection graphs, using the distance information between objects, to represent the spatial relationship among objects (Section \ref{subsec-construct-graphs}).
Second, we adopt an advanced common subgraph matching method inspired by \cite{Introduction-matching} to retrieve all potential common subgraphs between two connection graphs and obtain the optimal vertex correspondences from these common subgraphs (Section \ref{subsec-mcs-matching}).
In order to guarantee the precision, a normalized Euclidean distance dissimilarity-based approach \cite{Introduction-euclidean-similarity} is proposed to recognize the falsely-matched robots (Section \ref{subsec-mismatched-pair-identification}).
Finally, those robots, which are not estimated but within UWB communication range, are localized based on the well-matched robots and the relevant UWB ranging measurements (Section \ref{subsec-positioning-unmatched-robots}).
The main contributions of this paper are summarized as follows:

\begin{itemize}
  \item We propose a distributed relative position estimation method for a group of homogeneous robots, leveraging UWB ranging and LiDAR measurements. The identification and relative position estimation are formulated as a common subgraph matching problem.
  \item We present an approach to recognize the falsely-matched robots due to obstacle occlusion, then re-localize these robots using the well-matched robots and the relevant UWB ranging measurements.
  \item We conducted extensive experiments to validate the performance of the proposed approach. The results show that our approach can effectively localize the robots in both LOS and NLOS scenes with the assistance of UWB.
\end{itemize}

The remainder of the paper is organized as follows: the related works are introduced in Section \ref{sec-related-work}.
Section \ref{sec-system-overview} describes the proposed approach in detail.
The experiments and results are demonstrated in Section \ref{sec-experiments-results}.
Finally, the conclusion and future works are summarized in Section \ref{sec-conclusion}.

\begin{figure*}[b]
\centering
\includegraphics[width=140mm]{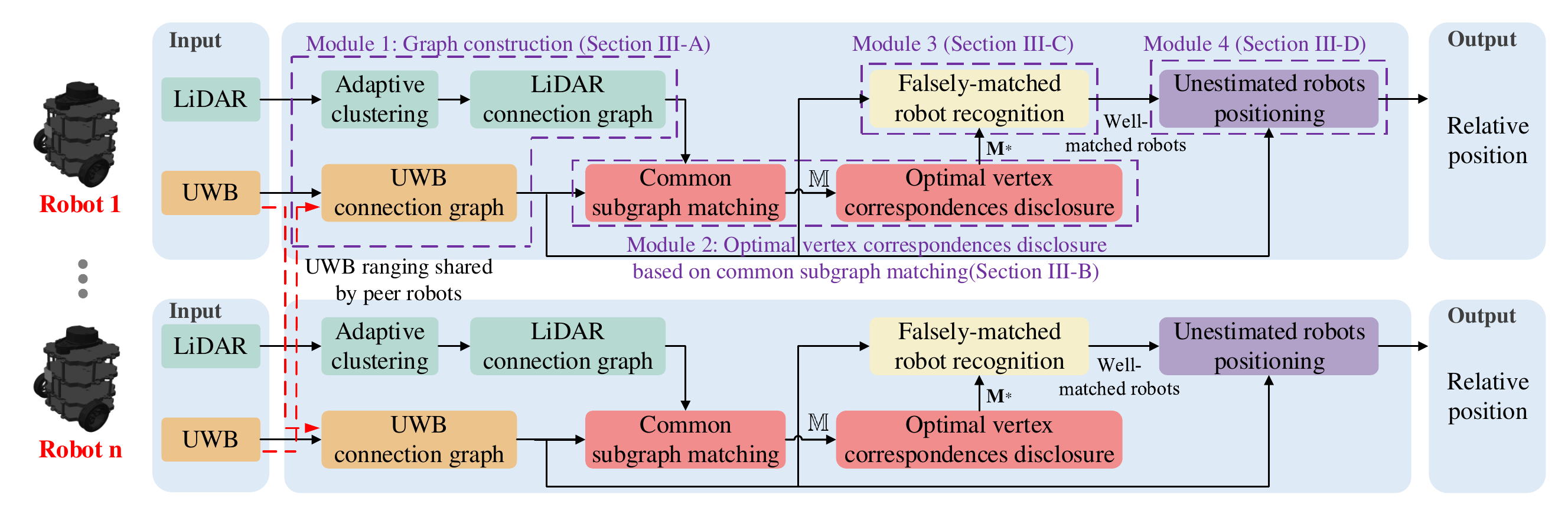}
\caption{The architecture of the proposed distributed relative localization approach.}
\label{figure-flowchart}
\end{figure*}

\section{Related work} \label{sec-related-work}
In the past decades, multi-robot systems have been widely adopted in various robotic applications, such as multi-robot exploration, formation, etc \cite{Related-work-exploration}\cite{Related-work-formation}.
Among these applications, the relative localization between robots is a fundamental component, allowing coordinated operations for multi-robot systems.
Visual, LiDAR, and UWB-based solutions are becoming popular, particularly in situations where GPS and communication are restricted \cite{Related-work-distributed-survey}.

For visual-based relative localization solutions, the classical methods require significant visual overlap between images from different robots \cite{Related-work-visual-01}\cite{Related-work-visual-02}.
Moreover, receiving the image or feature information from another robot depends on favorable communication conditions.
To reduce the reliance on communication, Pavliv et al. \cite{Related-work-visual-2} estimated the relative position of each robot based on a headset with a camera and IMU, employing the YOLOv3 algorithm \cite{Related-work-yolov3} and agent tracker. This method assigns ID information to each robot automatically, but the IDs will be reassigned randomly after being re-tracked.
To guarantee the data association, Xun et al. \cite{Related-work-visual-3} utilized active infrared LEDs to encode drone ID for easily identifying each drone. They then employed the Double Sphere (DS)
projection model of the fisheye camera \cite{Related-work-DS} and UWB ranging measurements to estimate relative position.
However, this method requires pre-designed passive LEDs with fixed flashing rates for ID detection.
It is not conducive to scalability for a multi-robot system.

In addition to visual-based solutions, LiDAR is also widely used in relative localization for multi-robot systems.
LiDAR-based clustering algorithms \cite{Introduction-adaptive-clustering} \cite{Related-work-lidar-clustering} are capable of clustering objects (including robots and obstacles) and estimating their positions in LiDAR's frame.
However, all clusters' IDs are designated randomly, resulting in challenges in data association and relative position estimation.
To address this issue, Zhu et al. \cite{Related-work-lidar-1} used a temporary tracker based on Kalman Filter (KF) to track robots and recognize the robot's ID by matching the trajectories of tracked objects and odometry from other robots.
However, this method requires that the robots move along non-parallel paths in favorable communication conditions.
Shalihan et al. \cite{Related-work-lidar-2} proposed an approach to identify robots from LiDAR clusters, using adaptive clustering \cite{Introduction-adaptive-clustering}, by comparing the UWB ranging with the cluster's distance. Nevertheless, this method lacks robustness and cannot localize the occluded robots.

While visual and LiDAR-based methods can provide excellent positioning accuracy, they are limited to  LOS scenes.
Radio-based relative localization methods, such as UWB sensors \cite{Related-work-radio} \cite{Related-work-uwb-1}, can work well in NLOS scenarios and provide unique identification for each robot.
Cao et al. \cite{Related-work-uwb-2} presented a relative localization method based on the least-square algorithm for a multi-robot system, which only relies on multiple UWB ranging nodes placed at different positions on the robot.
Despite its fully distributed nature, this method shows limited accuracy and is unsuitable for small robots.
To enhance the performance of UWB-based solutions, Liu et al. \cite{Related-work-uwb-3} and Nguyen et al. \cite{Related-work-uwb-4} accomplished robot-to-robot relative pose estimation based on optimization method, using a sequence of odometry and UWB ranging measurements collected continuously over the network.
Table \ref{table-related-work} presents a comprehensive survey of existing studies on relative localization.

\begin{table}[t]
\setlength{\abovecaptionskip}{5pt}%
\setlength{\belowcaptionskip}{5pt}%
\caption{Review of the related work on relative localization.}
\label{table-related-work}
\begin{threeparttable}
\resizebox{.49\textwidth}{!}{
\begin{tabular}{ccccc}
\hline
\multicolumn{1}{|c|}{\textbf{Method}} & \multicolumn{1}{c|}{\textbf{Platform}} & \multicolumn{1}{c|}{\textbf{Sensors}}                                               & \multicolumn{1}{c|}{\textbf{Data exchanged}}                                       & \multicolumn{1}{c|}{\textbf{Appraoch}}       \\ \hline \vspace{-8pt}
                                      &                                        &                                                                                     &                                                                                    &                                              \\ \hline
\multicolumn{1}{|c|}{\cite{Related-work-visual-01}}        & \multicolumn{1}{c|}{UGV }               & \multicolumn{1}{c|}{Visual}                                                         & \multicolumn{1}{c|}{3D points+Ctrl cmd.}                                               & \multicolumn{1}{c|}{EKF}                     \\ \hline
\multicolumn{1}{|c|}{\cite{Related-work-visual-02}}        & \multicolumn{1}{c|}{UAV  }               & \multicolumn{1}{c|}{Visual+IMU}                                                     & \multicolumn{1}{c|}{3D features+Odom.}                                             & \multicolumn{1}{c|}{EKF}                     \\ \hline
\multicolumn{1}{|c|}{\cite{Related-work-visual-3}}        & \multicolumn{1}{c|}{UGV+UAV   }               & \multicolumn{1}{c|}{\begin{tabular}[c]{@{}c@{}}Visual+IMU+\\ UWB+LEDs\end{tabular}} & \multicolumn{1}{c|}{\begin{tabular}[c]{@{}c@{}}Mutual meas.\\ +Odom.\end{tabular}} & \multicolumn{1}{c|}{ESKF}                    \\ \hline
\multicolumn{1}{|c|}{\cite{Related-work-visual-2}}        & \multicolumn{1}{c|}{UAV  }               & \multicolumn{1}{c|}{Visual+IMU}                                                     & \multicolumn{1}{c|}{N/A}                                                           & \multicolumn{1}{c|}{Machine learning}        \\ \hline
\multicolumn{1}{|c|}{\cite{Related-work-uwb-2}}        & \multicolumn{1}{c|}{UGV }               & \multicolumn{1}{c|}{Four UWBs}                                                   & \multicolumn{1}{c|}{N/A}                                                           & \multicolumn{1}{c|}{Non-linear optimization} \\ \hline
\multicolumn{1}{|c|}{\cite{Related-work-uwb-3}}        & \multicolumn{1}{c|}{UGV }               & \multicolumn{1}{c|}{Odom.+UWB}                                                      & \multicolumn{1}{c|}{Odom.}                                                         & \multicolumn{1}{c|}{Non-linear optimization} \\ \hline
\multicolumn{1}{|c|}{\cite{Related-work-uwb-4}}        & \multicolumn{1}{c|}{UAV  }               & \multicolumn{1}{c|}{Odom.+UWB}                                                      & \multicolumn{1}{c|}{Odom.}                                                         & \multicolumn{1}{c|}{Convex optimization}     \\ \hline
\multicolumn{1}{|c|}{\cite{Related-work-lidar-1}}        & \multicolumn{1}{c|}{UAV  }               & \multicolumn{1}{c|}{LiDAR+IMU}                                                      & \multicolumn{1}{c|}{Odom.}                                                         & \multicolumn{1}{c|}{Trajectory matching}     \\ \hline
\multicolumn{1}{|c|}{\cite{Related-work-lidar-2}}        & \multicolumn{1}{c|}{UGV }               & \multicolumn{1}{c|}{LiDAR+UWB}                                                      & \multicolumn{1}{c|}{N/A}                                                           & \multicolumn{1}{c|}{Distance matching}       \\ \hline
\multicolumn{1}{|c|}{\textbf{Ours}}   & \multicolumn{1}{c|}{UGV }               & \multicolumn{1}{c|}{LiDAR+UWB}                                                      & \multicolumn{1}{c|}{UWB ranging}                                                   & \multicolumn{1}{c|}{Subgraph matching}       \\ \hline
\end{tabular}
}
\scriptsize 
\begin{minipage}{.49\textwidth} 
\end{minipage}
\end{threeparttable}
\end{table}


\section{System Overview} \label{sec-system-overview}
In this section, we elaborate on the proposed distributed relative localization approach.
An overview of the proposed system is illustrated in Fig. \ref{figure-flowchart}, which consists of four modules: construct UWB connection graph and LiDAR connection graph using UWB ranging and LiDAR cluster information (Section \ref{subsec-construct-graphs}), determine the optimal vertex correspondences between two graphs based on common subgraph matching (Section \ref{subsec-mcs-matching}), recognize suspicious vertex correspondences induced by NLOS (Section \ref{subsec-mismatched-pair-identification}), localize robots that are not estimated using the well-matched robots and the UWB ranging measurements in the UWB connection graph (Section \ref{subsec-positioning-unmatched-robots}).

In order to better understand the proposed distributed relative localization system, the important notations used in this paper are listed in Table \ref{nomenclature-table}.

\begin{table}[t]
\centering
\setlength{\abovecaptionskip}{0pt}%
\setlength{\belowcaptionskip}{5pt}%
\caption{The critical notations used in this paper.}
\label{nomenclature-table}
\setlength{\tabcolsep}{1.1mm}{
\begin{tabular}{|c|c|}
\hline
\textbf{Notation}                 & \textbf{Description}                                                                                                                                                                                  \\ \hline
$N$                      & The number of robots in the team.                                                                                                                                                           \\ \hline
$\mathcal{G}_{uwb}$      & UWB connection graph.                                                                                                                                                                       \\ \hline
$\mathcal{G}_{lidar}$    & LiDAR connection graph.                                                                                                                                                                     \\ \hline
$L(v_{i})$               & Label sequence of vertex $v_{i}$ in $\mathcal{G}_{uwb}$.                                                                                                                                    \\ \hline
$L(u_{l})$               & Label sequence of vertex $u_{l}$ in $\mathcal{G}_{lidar}$.                                                                                                                                  \\ \hline
$\textbf{L}_{uwb}$       & Set of label sequences of unmatched vertices in $\mathcal{G}_{uwb}$.                                                                                                                              \\ \hline
$\textbf{L}_{lidar}$     & Set of label sequences of unmatched vertices in $\mathcal{G}_{lidar}$.                                                                                                                            \\ \hline
$S$                      & Set of the unmatched vertices in $\mathcal{G}_{uwb}$ and $\mathcal{G}_{lidar}$.                                                                                                                            \\ \hline

$r_{i,j}$                & UWB ranging between Robot$i$ and Robot$j$, $i\neq j$.                                                                                                                                       \\ \hline
$\textbf{r}_{i}$         & Set of all UWB ranging data measured by Robot$i$.                                                                                                                                           \\ \hline
$\textbf{R}_{i}$         & \begin{tabular}[c]{@{}c@{}}Set of all UWB ranging data collected by Robot$i$, which\\ consists of the UWB ranging data measured by itself and\\ shared by surrounding robots.\end{tabular}    \\ \hline
$d_{l,k}$                & \begin{tabular}[c]{@{}c@{}}Distance between vertex $u_{l}$ and vertex $u_{k}$ in $\mathcal{G}_{lidar}$.\end{tabular}                                                                      \\ \hline
$\bar{\textbf{x}}_{i,j}$ & \begin{tabular}[c]{@{}c@{}}Relative position of Robot$j$ in Robot$i$'s local frame.\end{tabular}                                                             \\ \hline
$\textbf{M}$             & A common subgraph between two graphs.                                                                                                                                                       \\ \hline
$\textbf{M}^{*}$         & \begin{tabular}[c]{@{}c@{}}Maximum common subgraph with highest similarity. \end{tabular}                                                                                                                         \\ \hline
$\mathbb{M}$             & Set of all common subgraphs between two graphs.                                                                                                                                             \\ \hline
$\mathbb{M}'$            & \begin{tabular}[c]{@{}c@{}}Set of common subgraphs with the largest size. \end{tabular}                                                                                                                         \\ \hline
$\lambda$                & \begin{tabular}[c]{@{}c@{}}Distance threshold used in common subgraph matching\\ part of \textit{Module 2} (Section \ref{retrieve common subgraph}).\end{tabular}                                    \\ \hline
$\epsilon$               & \begin{tabular}[c]{@{}c@{}}Dissimilarity threshold used in \textit{Module 3} (Section \ref{subsec-mismatched-pair-identification}).\end{tabular} \\ \hline
$\textbf{O}$            & \begin{tabular}[c]{@{}c@{}}Set of vertex correspondences for vertices that are not in $\textbf{M}^{*}$. \end{tabular}                                                                                                                         \\ \hline
$\textbf{Q}$             & \begin{tabular}[c]{@{}c@{}}Set of optimal vertex correspondences, including $\textbf{M}^{*}$ and $\textbf{O}$.\end{tabular}                 \\ \hline
\end{tabular}}
\end{table}

\subsection{Module 1: graph construction} \label{subsec-construct-graphs}
\begin{figure}[t]
\centering
\subfloat[UWB connection graph]{
    \includegraphics[width=42mm]{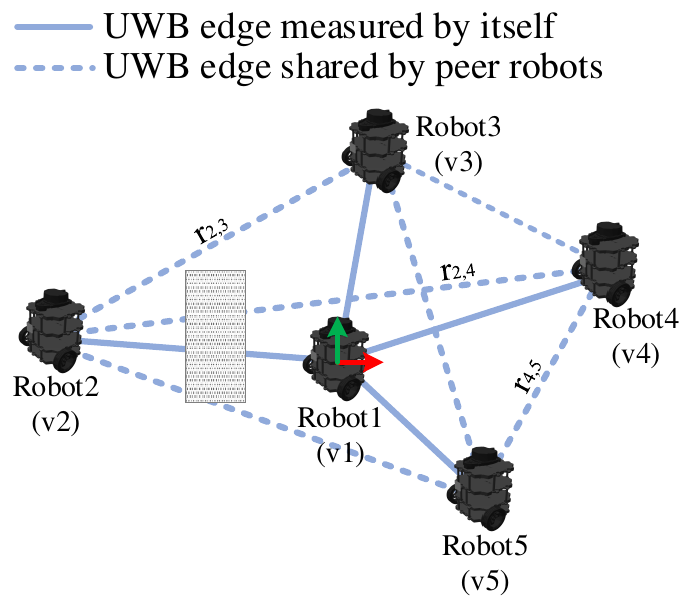}
    \label{construct_graph-uwb}
}
\subfloat[LiDAR connection graph]{
    \includegraphics[width=42mm]{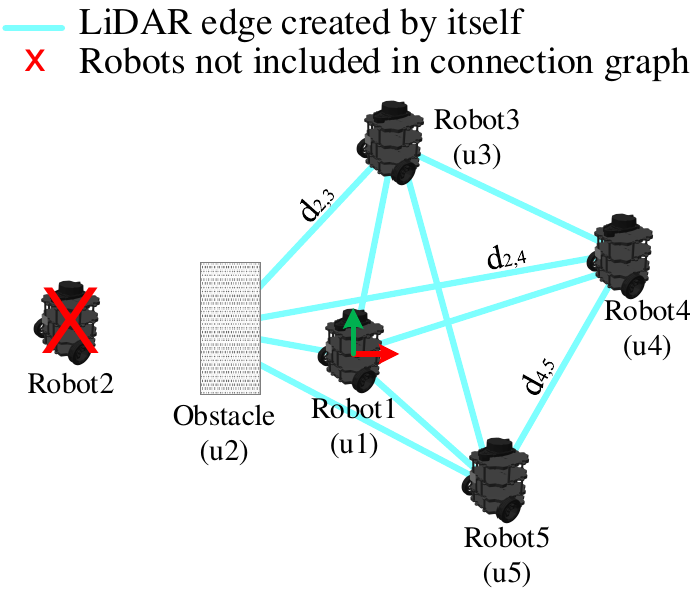}
    \label{construct_graph-lidar}
}
\caption{The illustration of the UWB and LiDAR connection graphs. (a) constructed by utilizing UWB ranging directly measured by itself (blue solid line) and those shared by peer robots (blue dotted line) based on the customized UWB communication protocol, as depicted in Section \ref{subsection-metrics-and-protocol}. (b) constructed by using the distance between each pair of clusters (aqua solid line) generated by the adaptive clustering algorithm \cite{Introduction-adaptive-clustering}. The robot that is not detected by LiDAR will not be considered as a vertex.}
\label{figure-construct-graph}
\end{figure}

In this module, we introduce the procedure of constructing connection graphs.
Fig. \ref{figure-construct-graph} illustrates building the UWB connection graph and LiDAR connection graph, in which the spatial relationship between vertices is presented with the distance value measured by UWB and LiDAR, respectively.
Each robot independently constructs its own connection graphs on the onboard computer.

In particular, for the UWB connection graph, the UWB sensor is able to measure the distance to surrounding robots and broadcast a limited amount of data to team members within the communication range.
For example, Robot1 collects UWB ranging set $\textbf{R}_{1}$, which is used to construct its UWB connection graph $\mathcal{G}_{uwb}=(V_{uwb}, E_{uwb})$.
Each vertex in the set $V_{uwb}=\left \{ v_{1}, v_{2}, ..., v_{n} \right \}$, $n\in N$, denotes one unique robot, in which the ID of the vertex corresponds to the robot ID and UWB ID.
The set of the UWB ranging between vertices is defined as the edge set $E_{uwb}=\left \{ r_{i,j} \right \}$, $r_{i,j} \in  \textbf{R}_{1}$.
In this paper, we consider the undirected connection graphs.
Hence, one edge could be represented by bidirectional UWB ranging measurements (e.g., both $r_{1,2}$ and $r_{2,1}$ can denote the edge between vertex $v_{1}$ and vertex $v_{2}$).

For the LiDAR connection graph, the adaptive clustering algorithm \cite{Introduction-adaptive-clustering} is employed to cluster objects (including robots and obstacles) and obtain the position of the objects in the local robot frame.
Some robots occluded by obstacles or out of scanning range cannot be detected via LiDAR.
Furthermore, we remove clusters with large sizes, as they significantly differ from the characteristics of robots.
The remaining clusters are utilized to construct the LiDAR connection graph $\mathcal{G}_{lidar}=(U_{lidar}, E_{lidar})$.
$U_{lidar}=\left \{ u_{1}, u_{2}, ..., u_{m} \right \}$ denotes the set of vertices in $\mathcal{G}_{lidar}$, containing robots and obstacles in FOV of LiDAR.
The vertex $u_{1}$ is always defined as the robot performing localization, while the vertex IDs for other clusters are assigned randomly.
We denote $d_{l,k}$ as the distance value between vertex $u_{l}$ and $u_{k}$, which is calculated based on the positions of LiDAR clusters.
In $\mathcal{G}_{lidar}$, each pair of vertices will be connected.
The edge set is represented as $E_{lidar}=\left \{ d_{l,k} \right \}$, $1\leq l,k\leq m$ and $l\neq k$.

Thanks to the characteristics of electromagnetic waves, UWB signals can penetrate certain obstacles.
However, LiDAR can only detect objects within LOS.
Additionally, some edges may be missing in $\mathcal{G}_{uwb}$ due to the communication constraints.
Thus, $\mathcal{G}_{uwb}$ and $\mathcal{G}_{lidar}$ may not be identical.
These two graphs will serve as input for the common subgraph matching part in the subsequent subsection.

\subsection{Module 2: optimal vertex correspondences disclosure based on common subgraph matching} \label{subsec-mcs-matching}
This module aims to figure out the vertex correspondences between the UWB connection graph and LiDAR connection graph, which contains two main components: 1) \textit{common subgraph matching} (Section \ref{retrieve common subgraph}) and 2) \textit{optimal vertex correspondences disclosure} (Section \ref{obtain optimal vertex correspondences}).
The former introduces a common subgraph matching-based approach to find all common subgraphs $\mathbb{M}$ between two graphs.
In this paper, we consider a common subgraph that includes a subset of vertices and all edges in the graphs that connect these vertex pairs in the subset, i.e., common induced subgraph \cite{graph-theory}.
Due to obstacle blockage and limited communication, a common subgraph may only involve a portion of vertices.
The latter is in charge of disclosing the set of optimal vertex correspondences $\textbf{Q}$ from $\mathbb{M}$, which includes vertex correspondences in the maximum common subgraph with the highest similarity $\textbf{M}^{*}$ and the scattered vertex correspondences $\textbf{O}$ obtained from the remaining common subgraphs.
The optimal vertex correspondences help in identifying the robots detected by LiDAR.

\begin{myDef}
    The vertex that has not been matched is referred to as an "unmatched vertex" during the matching process.
\end{myDef}

\subsubsection{Common subgraph matching} \label{retrieve common subgraph}
A common subgraph is made up of several vertex correspondences $\textbf{M}=\left \{ \textbf{m}_{i,l} \right \}$, where $\textbf{m}_{i,l}=(v_{i}, u_{l})$, $v_{i}\in V_{uwb}$ and $u_{l}\in U_{lidar}$.
Each unmatched vertex in $\mathcal{G}_{uwb}$ and $\mathcal{G}_{lidar}$ carries a label sequence, i.e., $\textbf{L}_{uwb}$ for $\mathcal{G}_{uwb}$ and $\textbf{L}_{lidar}$ for $\mathcal{G}_{lidar}$, for determining whether two vertices are matched.
This part aims to find all common subgraphs $\mathbb{M}$ between $\mathcal{G}_{uwb}$ and $\mathcal{G}_{lidar}$, which consists of four steps: \textbf{Initialize}, \textbf{Extend}, \textbf{Update}, and \textbf{Backtrack}.
The first step is to initialize $\textbf{M}$, $\textbf{L}_{uwb}$, and $\textbf{L}_{lidar}$ for a recursive function $\textit{Retrieve}$ operation.
The second and third steps are jointly performed in each recursion to find the common subgraph.
Once a common subgraph is found, the fourth step incorporates $\textbf{M}$ into $\mathbb{M}$ and backtracks the $\textit{Retrieve}$ function to find the next common subgraph continuously.

\textbf{Initialize:} We build up a common subgraph $\textbf{M}$ using the depth-first search, commencing with a known vertex correspondence $\left \{ (v_{1},u_{1}) \right \}$.
This is because vertex $v_{1}$ and $u_{1}$ in $\mathcal{G}_{uwb}$ and $\mathcal{G}_{lidar}$ both represent the robot performing localization.
With this known association, the computational time for finding the common subgraph will be reduced.
Then, we initialize the label sequence of each unmatched vertex in the two graphs based on whether it is adjacent to $v_{1}$ or $u_{1}$.
If unmatched vertices in $\mathcal{G}_{uwb}$ are adjacent to $v_{1}$, their label sequences will be updated with $L(v_{i})=L(v_{i})\cup r_{1,i}$; otherwise, $L(v_{i})=L(v_{i})\cup \infty$, where $v_{i}\in V_{uwb}$.
Likewise, $L(u_{l})=L(u_{l})\cup d_{1,l}$ if $u_{l}$ is adjacent to $u_{1}$; otherwise, $L(u_{l})=L(u_{l})\cup \infty$, where $u_{l}\in U_{lidar}$.

\begin{myDef}
    Function $f(A, B, \lambda)$ is defined to compare if the absolute difference between all elements in sequence $A=\left \{ a_{1}, a_{2},\cdots, a_{n}\right \}$ and sequence $B=\left \{ b_{1}, b_{2},\cdots, b_{n}\right \}$ is less than the given $\lambda$, i.e., $\left | a_{i}-b_{i} \right |< \lambda, \forall i\in \left \{ 1,2,\cdots,n \right \}$. If true, $f(A, B, \lambda)=1$; otherwise, $f(A, B, \lambda)=0$. 
\end{myDef}

\textbf{Extend:} Once the \textbf{Initialize} step is done, the function \textit{Retrieve} is performed to find all common subgraphs $\mathbb{M}$ in a recursive manner.
To find a common subgraph, we extend $\textbf{M}$ in each recursion by a vertex correspondence $\left \{ (v_{i},u_{l}) \right \}$ if the absolute difference between the corresponding elements in label sequences $L(v_{i})$ and $L(u_{l})$ is less than a distance threshold $\lambda$ (see line 10-11 in Algorithm \ref{graph_matching_algo}):
\begin{equation}
 \textbf{M}=\left\{\begin{matrix}
\textbf{M}\cup \left \{ (v_{i}, u_{l}) \right \}, & \text{if\ } f(L(v_{i}), L(u_{l}), \lambda)=1  \\
 \textbf{M}, & \text{otherwise} \\
\end{matrix}\right.
\end{equation}

\textbf{Update:} After $\textbf{M}$ is extended in each recursion, the two graphs and label sequence sets will be updated.
Specifically, the latest matched vertices ($v_{i}$ and $u_{l}$) and their corresponding label sequences ($L(v_{i})$ and $L(u_{l})$) will be removed.
Then, the label sequence of each unmatched vertex in two graphs is updated according to the rule mentioned in the \textbf{Initialize} step, depending on whether they are adjacent to the vertices $v_{i}$ or $u_{l}$.
The updated graphs and label sequence sets are denoted as $\mathcal{G}'_{uwb}$, $\mathcal{G}'_{lidar}$, $\textbf{L}'_{uwb}$, and $\textbf{L}'_{lidar}$, which will work as new input for the recursive call of the \textit{Retrieve} function (see line 13 in Algorithm \ref{graph_matching_algo}).

\textbf{Backtrack:} Once a common subgraph $\textbf{M}$ is found,  it will be incorporated into $\mathbb{M}$, $\mathbb{M}\leftarrow \mathbb{M}\cup \textbf{M}$ (see line 5 in Algorithm \ref{graph_matching_algo}).
Meanwhile, we backtrack to find the next common subgraph.
Specifically, the vertex correspondence that was newly added to $\textbf{M}$ will be removed, $\textbf{M}\leftarrow \textbf{M}\setminus  \left \{ (v_{i}, u_{l}) \right \}$ (see line 14 in Algorithm \ref{graph_matching_algo}).

A variable $bound$ is used to effectively determine if a common subgraph has been found.
It is computed based on the label sequences of the vertices that remain unmatched in two graphs.
A common subgraph is found when the value of $bound$ is 0, indicating that no additional vertex correspondences can be added to the current common subgraph $\textbf{M}$.

\begin{equation}
\begin{split}
    &bound=\sum_{s\in S}\min(f(L(v_{i}), L(s), \; \lambda), \  
    f(L(u_{l}), L(s), \; \lambda)), \\
&\forall v_{i} \in V_{uwb} \  \text{and} \  \forall u_{l} \in U_{lidar},
\end{split}
\label{equation-bound}
\end{equation}
where $S$ is the set of the current unmatched vertices in the UWB connection graph and LiDAR connection graph.

\subsubsection{Optimal vertex correspondences disclosure}\label{obtain optimal vertex correspondences}
Here, we proceed to discover the set of optimal vertex correspondences $\textbf{Q}$ from $\mathbb{M}$ once all common subgraphs are found.
$\textbf{Q}$ consists of two parts: $\textbf{M}^{*}$ and $\textbf{O}$.
$\textbf{M}^{*}$ is the maximum common subgraph with the highest similarity in $\mathbb{M}$, but it may only include partial vertices due to obstacle occlusion and limited communication conditions.
$\textbf{O}$ represents the set of vertex correspondences for those vertices that are not in $\textbf{M}^{*}$.

In general, a common subgraph with a larger size can provide better discrimination.
Thus, we try to find the maximum common subgraph from $\mathbb{M}$ based on the size of each common subgraph.
However, multiple common subgraphs with similar sizes may exist.
The set of these common subgraphs with the same size is defined as $\mathbb{M}'$.
In order to recognize the best maximum common subgraph, each common subgraph is evaluated using a dissimilarity score derived from the normalized Euclidean distance.
According to Equation 4, a lower value indicates a higher similarity.
Hence, we take the common subgraph with the lowest dissimilarity value into account when they have the same size.
Such a rule can be written as:
\begin{equation}
\textbf{M}^{*}=\mathop{\text{arg}\min}\limits_{\textbf{M}\in\mathbb{M}'}dissimilarity(\textbf{M}),
\label{equation-maximum-subgraph}
\end{equation}  
where function $dissimilarity(\textbf{M})$ is defined as follows:

\begin{equation}
\setlength{\abovedisplayskip}{0pt}%
dissimilarity(\textbf{M})=\frac{1}{size(\textbf{M})}\sqrt{\sum\limits_{\textbf{m}_{i,l}=(v_{i},u_{l})\in \textbf{M}, \atop \textbf{m}_{j,k}=(v_{j},u_{k})\in \textbf{M}} (r_{i,j}-d_{l,k})^2},
\end{equation}  
where $dissimilarity(.)$ stands for the common subgraph's dissimilarity score.
$size(.)$ is the size of the common subgraph.

The correspondences for those vertices that have already been matched but are not included in $\textbf{M}^{*}$ can still provide valuable references.
Therefore, following a similar rule mentioned in Section \ref{obtain optimal vertex correspondences} above, we endeavor to disclose vertex correspondences for these vertices from the remaining common subgraphs.
The set of these scattered vertex correspondences is denoted as $\textbf{O}$.
Finally, the set of the optimal vertex correspondences can be obtained, $\textbf{Q}=\left \{ \textbf{M}^{*}, \textbf{O} \right \}$.
The overview of the optimal vertex correspondences disclosure based on common subgraph matching module is depicted in Algorithm \ref{graph_matching_algo}.

\begin{algorithm}[t]
\caption{Optimal vertex correspondences disclosure based on common subgraph matching}
\label{graph_matching_algo}
\SetKwData{In}{\textbf{in}}\SetKwData{To}{to}
\DontPrintSemicolon
\SetAlgoLined
\KwIn {UWB connection graph $\mathcal{G}_{uwb}$ \quad \quad \quad \quad \quad
LiDAR connection graph $\mathcal{G}_{lidar}$}
\KwOut {Optimal vertex correspondence set $\textbf{Q}$}

Initialize $\textbf{M}$, $\textbf{L}_{uwb}$, and $\textbf{L}_{lidar}$ based on the rule described in \textbf{Initialize}, and $\mathbb{M}\leftarrow \varnothing $; 

\SetKwFunction{Retrieve}{Retrieve}
\SetKwProg{Fn}{Function}{:}{}
\Fn{\Retrieve{\rm $\mathcal{G}_{uwb}$,$\mathcal{G}_{lidar}$,$\textbf{L}_{uwb}$,$\textbf{L}_{lidar}$,$\textbf{M}$}}
{
 Calculate \textit{bound} based on Equation 2; \\
 \tcp*[h]{\small a common subgraph {\rm \textbf{M}} is found} \\
\If{$\textit{bound}=0$}{
      $\mathbb{M}\leftarrow \mathbb{M}\cup \textbf{M}$;   \\
      \textbf{return}; 
    }

\For{$v_{i}\in V_{uwb}$}{
    \For{$u_{l}\in U_{lidar}$}{
    \If{$f(L(v_{i}), L(u_{l}), \lambda) = 1$}{
      \tcp*[h]{\small extend {\rm \textbf{M}} using $\left \{ (v_{i}, u_{l}) \right \}$} \\ 
      $\textbf{M}\leftarrow \textbf{M}\cup \left \{ (v_{i}, u_{l}) \right \}$;   \\
      Update $\mathcal{G}_{uwb}$, $\mathcal{G}_{lidar}$, $\textbf{L}_{uwb}$, and $\textbf{L}_{lidar}$ based on the rule described in \textbf{Update};    \\
      \Retrieve{\rm $\mathcal{G}'_{uwb}$,$\mathcal{G}'_{lidar}$,$\textbf{L}'_{uwb}$,$\textbf{L}'_{lidar}$,$\textbf{M}$};    \\ 
      \tcp*[h]{\small remove the latest vertex correspondence $\left \{ (v_{i}, u_{l}) \right \}$ from {\rm \textbf{M}} for the next common subgraph} \\
      $\textbf{M}\leftarrow \textbf{M}\setminus  \left \{ (v_{i}, u_{l}) \right \}$;
    }
    }
}
}

Disclose the set of optimal vertex correspondences $\textbf{Q}=$ $\left \{ \textbf{M}^{*}, \textbf{O} \right \}$ from $\mathbb{M}$, following the rule mentioned in Section \ref{obtain optimal vertex correspondences}; \\
\quad \textbf{return} $\textbf{Q}$
\end{algorithm}

\subsection{Module 3: falsely-matched robot recognition} \label{subsec-mismatched-pair-identification}
The correspondence between robots and LiDAR clusters can be figured out by the optimal vertex correspondence set $\textbf{Q}$ obtained from \textit{Module 2}.
Therefore, we can identify the robots from a set of anonymous LiDAR clusters and obtain the relative position of the identified robots in the local robot frame according to the corresponding LiDAR cluster information.
However, a few falsely-matched correspondences may occur due to obstacle occlusion and constrained communication.
To tackle the falsely-matched issue, we present an approach to verify the matched robots based on the robots' position and relevant UWB ranging measurements, thereby ensuring the system's accuracy.

The maximum common subgraph with the highest similarity, $\textbf{M}^{*}$, exhibits the most excellent reliability.
Thus, we assume that these robots in $\textbf{M}^{*}$ are correctly matched and consider them as the benchmark for verifying the robots in $\textbf{O}$.
For example, as shown in Fig. \ref{figure-construct-graph}, we assume that the robot to be verified is Robot2, and the benchmark robots are Robot1, Robot3, Robot4, and Robot5.
The distances between Robot2 and the robots in the benchmark can be calculated according to their position information.
These distance values are organized into a distance vector.
Then, the relevant UWB ranging measurements between robots in the UWB connection graph, i.e., $r_{1,2}$, $r_{2,3}$, $r_{2, 4}$, and $r_{2, 5}$, form another distance vector.
Finally, we compute the normalized Euclidean distance dissimilarity between these two distance vectors based on Equation 4.
If the value is higher than a dissimilarity threshold $\epsilon $, the robot will be determined as falsely-matched.

\begin{myDef}
    The robots that are not matched by Module 2 or recognized as falsely-matched by Module 3 are referred to as "unestimated robots".
\end{myDef}

\subsection{Module 4: unestimated robots positioning} \label{subsec-positioning-unmatched-robots}
Based on the modules depicted in Section \ref{subsec-mcs-matching} and Section \ref{subsec-mismatched-pair-identification}, the relative positions of most robots in LOS scenarios can be accurately estimated. 
Nevertheless, the position information of those unestimated robots remains crucial in a multi-robot system, especially for a cooperative exploration task.
In this paper, the well-matched robots and the corresponding UWB ranging measurements are utilized to localize those unestimated robots within the UWB communication range.
The well-matched robots, as verified by \textit{Module 3} in Section \ref{subsec-mismatched-pair-identification}, serve as the anchors for the localization of unmatched robots.
The position of unestimated Robot$j$ in the Robot$i$'s local frame can be estimated by minimizing the residual error of the UWB ranging measurements, which is formulated as follows:
\begin{equation} \label{least_square}
    \bar{\textbf{x}}_{i ,j }=\mathop{\arg\min}\limits_{\textbf{x}_{i,j}}\sum_{k\in K}(\underbrace{r_{k,j}}_{\mathclap{uwb \atop ranging}}-d(\textbf{x}_{i,j}, \underbrace{\bar{\textbf{x}}_{i,k}}_{\mathclap{well-matched \atop robots}}))^{2},
\end{equation}
where function $d(.)$ computes the distance value between the known position of Robot$k$ and the given position $\textbf{x}_{i,j}$.
$K$ is the set of well-matched robots, which should be greater than three.
The algorithm turns out to find the best position configuration to satisfy the UWB ranging constraints based on non-linear least-square optimization \cite{system-max-likelihood}.


\section{Experimental results} \label{sec-experiments-results}
In this section, we are going to validate the feasibility, reliability, and scalability of the proposed distributed relative localization approach using extensive experiments with a group of homogeneous robots in communication-limited scenes.
Section \ref{subsection-metrics-and-protocol} introduces the evaluation metrics and UWB communication protocol adopted in our approach.
We disclose the performance of different modules proposed in Section \ref{subsection-evaluation-different-modules}.
Subsequently, the performance of the proposed approach under different communication conditions is discussed in Section \ref{impact communication}.
Section \ref{subsection-different-sorting-methods} highlights the significance of different communication protocols for sharing UWB ranging.
Section \ref{robots scale evaluation} studies the scalability with varying numbers of robots. 
Finally, the proposed approach is also evaluated in a simple dynamic scenario in Section \ref{evaluation dynamic scenarios}.

\subsection{Evaluation metrics and UWB communication protocol}
\label{subsection-metrics-and-protocol}
To better evaluate the performance of the proposed approach, we employ two metrics, namely relative position error and solution rate.
These two metrics can provide a quantitative measure for assessing our approach, as below:
\begin{enumerate}
    \item \textbf{Relative position error (meter):} The relative position error is used to evaluate the accuracy of the proposed approach. In this paper, we consider the average value of all the results for the robot at different timestamps.

    \item \textbf{Solution rate (\%):} The solution rate is utilized to evaluate the effectiveness of the proposed approach, which is a ratio between the number of successful solutions and the total number of trials made (around 130 trials).
    The solution rate of 0\% means that the robot cannot be localized, whereas that of 100\% signifies that the robot can always be localized in every trial.
\end{enumerate}

\begin{figure}[t]
\centering
\includegraphics[width=90mm]{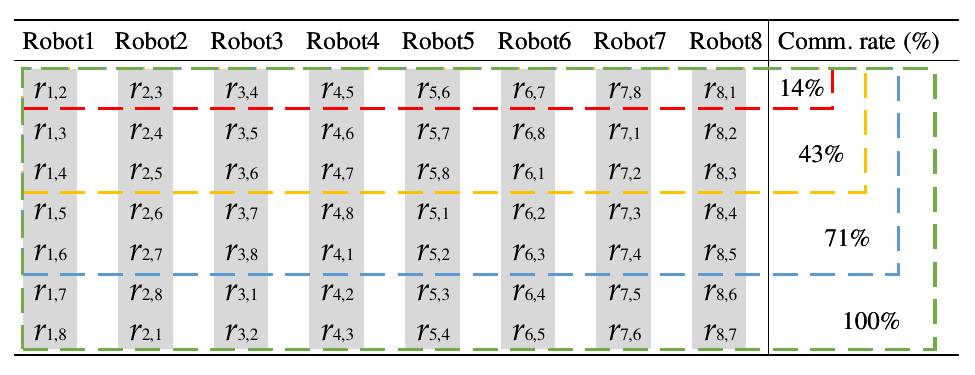}
\caption{The UWB communication protocol for different communication rates.}
\label{figure-communication-protocol}
\end{figure}

Furthermore, considering communication resource limitations, each robot only shares a portion of its UWB ranging data with nearby robots.
In order to ensure fairness analysis and maximize the connectivity of the UWB connection graph, we specify a simple UWB communication protocol for each robot, as shown in Fig. \ref{figure-communication-protocol}.
The UWB ranging data measured by each robot are arranged in a fixed order.
The communication rate is defined as:
\begin{equation}
    \text{communication rate} = \frac{N_{\text{shared}}}{N_{\text{total}}}\times 100\%,
\end{equation}
where $N_{\text{shared}}$ denotes the number of UWB ranging data shared by each robot, and $N_{\text{total}}$ represents the total number of UWB ranging data measured by each robot, which is seven according to our experimental setup.

\begin{figure}[t]
\centering
\includegraphics[width=90mm]{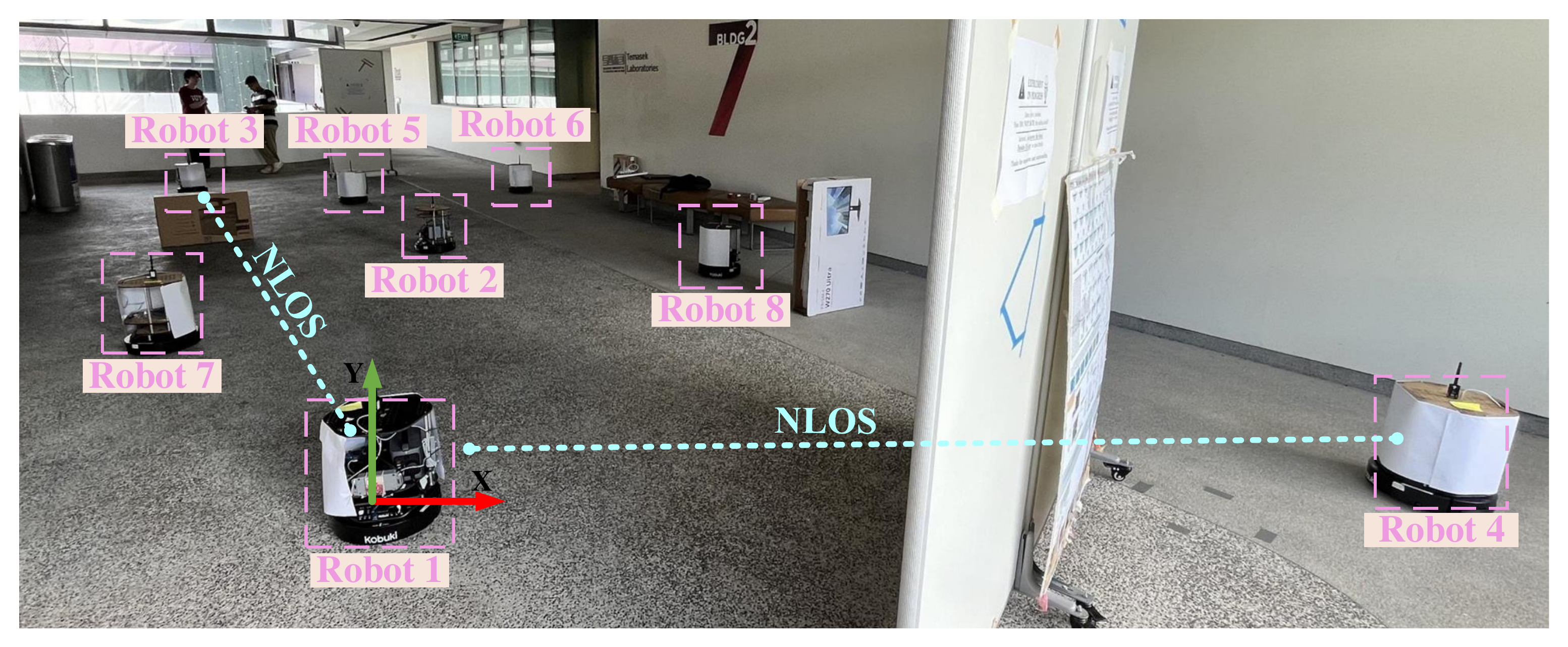}
\caption{The experimental setups in an indoor environment with 6m$\times$17m, including 8 robots and several obstacles. Robot 1 is NLOS to Robot 3 and Robot 4 (aqua dotted line). Note that paper wrapping only provides a controllable LiDAR detection for performance evaluation.}
\label{figure-real-world-experiment}
\end{figure}

\subsection{Evaluation of different modules}
\label{subsection-evaluation-different-modules}
In this subsection, we conduct a representative experiment to evaluate the designed modules in a real-world environment.
As shown in Fig. \ref{figure-real-world-experiment}, the experiment was performed in an indoor environment with an area of 6m$\times$17m, where 8 Turtlebot2 robots and several static obstacles (including boxes and partitions) were arbitrarily placed.
Each robot carries a 2D LiDAR (RPLIDAR A3, maximum range of 25m with 360$^{\circ}$ FOV) and a UWB sensor with a unique ID (Nooploop Linktrack, maximum range of 100m, sampling frequency of 50Hz).
The data collection and relative position estimation were independently run on each robot's local computer, with an Intel(R) Xeon(R) W-10855M CPU, a frequency of 2.80GHz, and a RAM of 32GB.
All programs were implemented on the Robot-Operating-System (ROS) \cite{experiment-ros}.
To facilitate readability, we analyze the performance of the proposed approach from Robot1's perspective, which is NLOS to Robot3 and Robot4.
Besides, it is worth noting that the adaptive clustering algorithm used in \textit{Module 1} takes approximately 12ms to generate LiDAR clusters from a LiDAR scan with 1300 points.

\vspace{6pt}
\subsubsection{Optimal vertex correspondences disclosure
based on common subgraph matching module (Module 2)}
\label{impact-distance}
Here, we discuss the performance of the optimal vertex correspondences disclosure based on common subgraph matching module depicted in Section \ref{subsec-mcs-matching} with different distance threshold settings $\lambda$.
For clarity, we set the communication rate to 100\%.
The experimental results of this module with different communication rates will be demonstrated in Section \ref{impact communication}.

\begin{figure}[h]
\centering
\subfloat[$\lambda $ = 0.2m]{
    \includegraphics[width=43mm]{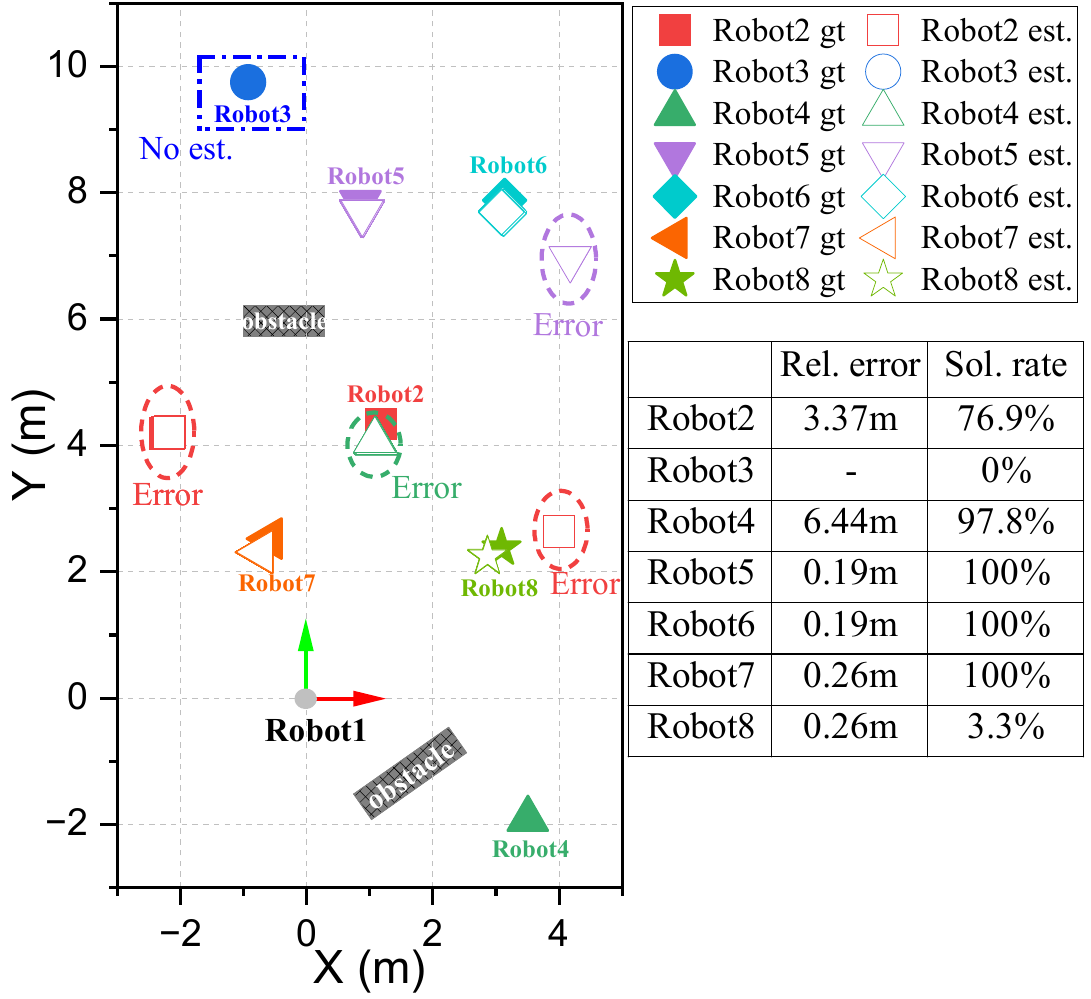}
    \label{distance-threshold-0.2}
}
\subfloat[$\lambda $ = 0.5m]{
    \includegraphics[width=43mm]{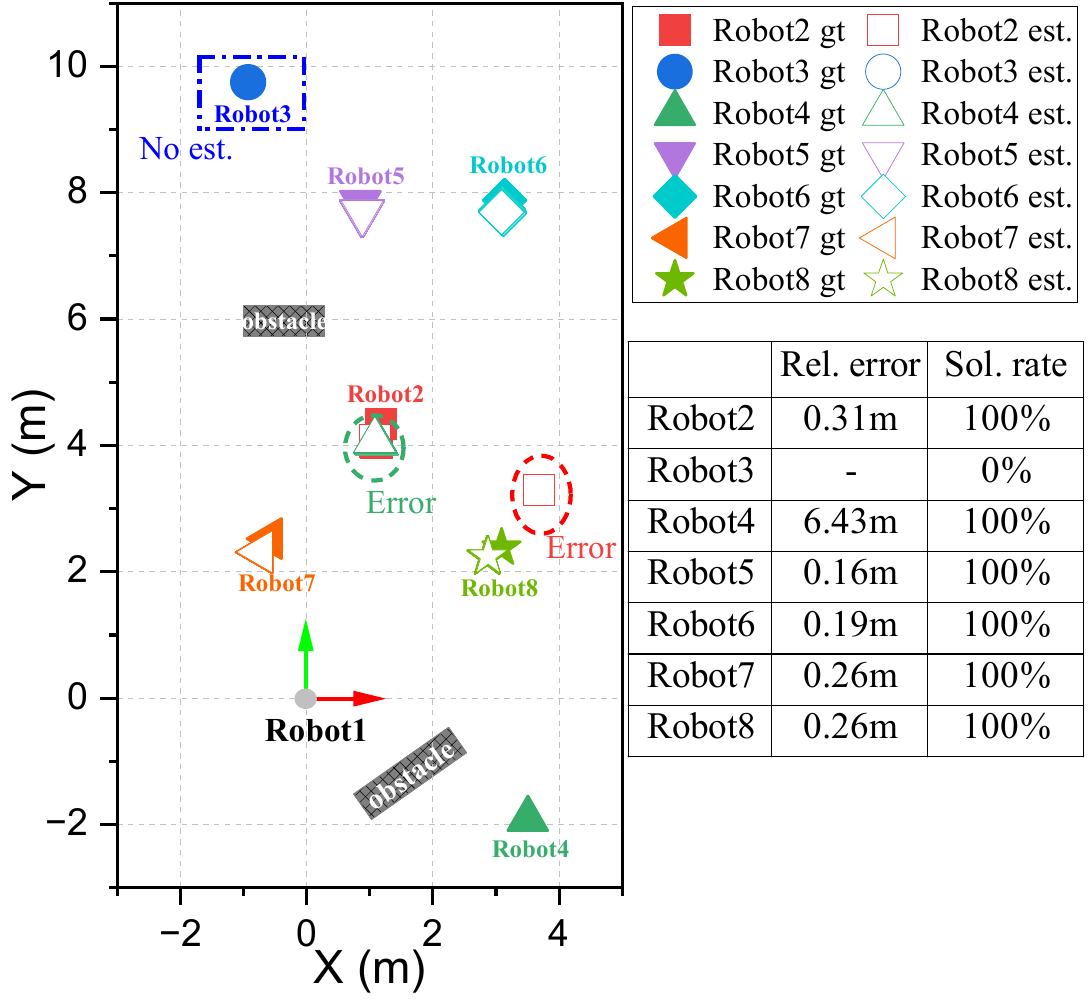}
    \label{distance-threshold-0.5}
}
\hfil
\subfloat[$\lambda $ = 1.0m]{
    \includegraphics[width=43mm]{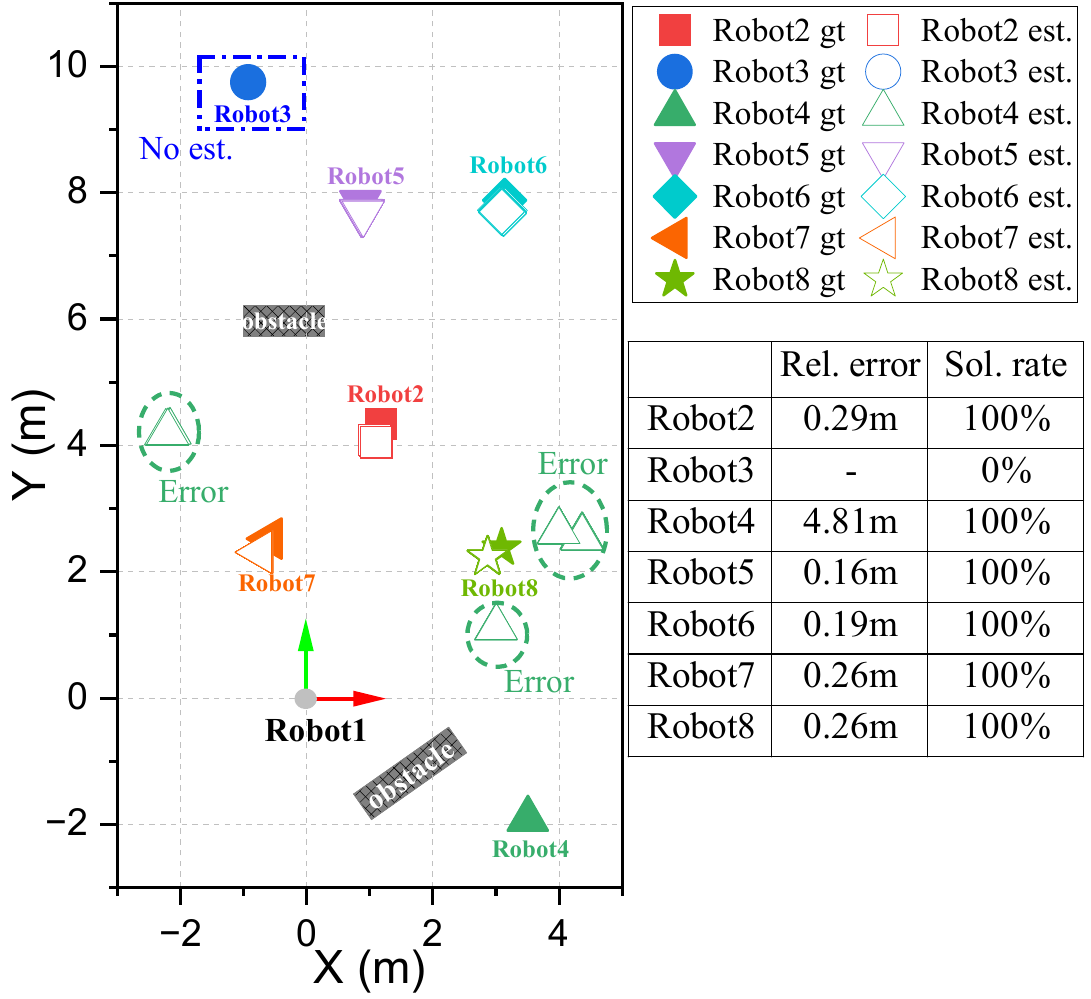}
    \label{distance-threshold-1.0}
}
\subfloat[$\lambda $ = 1.5m]{
    \includegraphics[width=43mm]{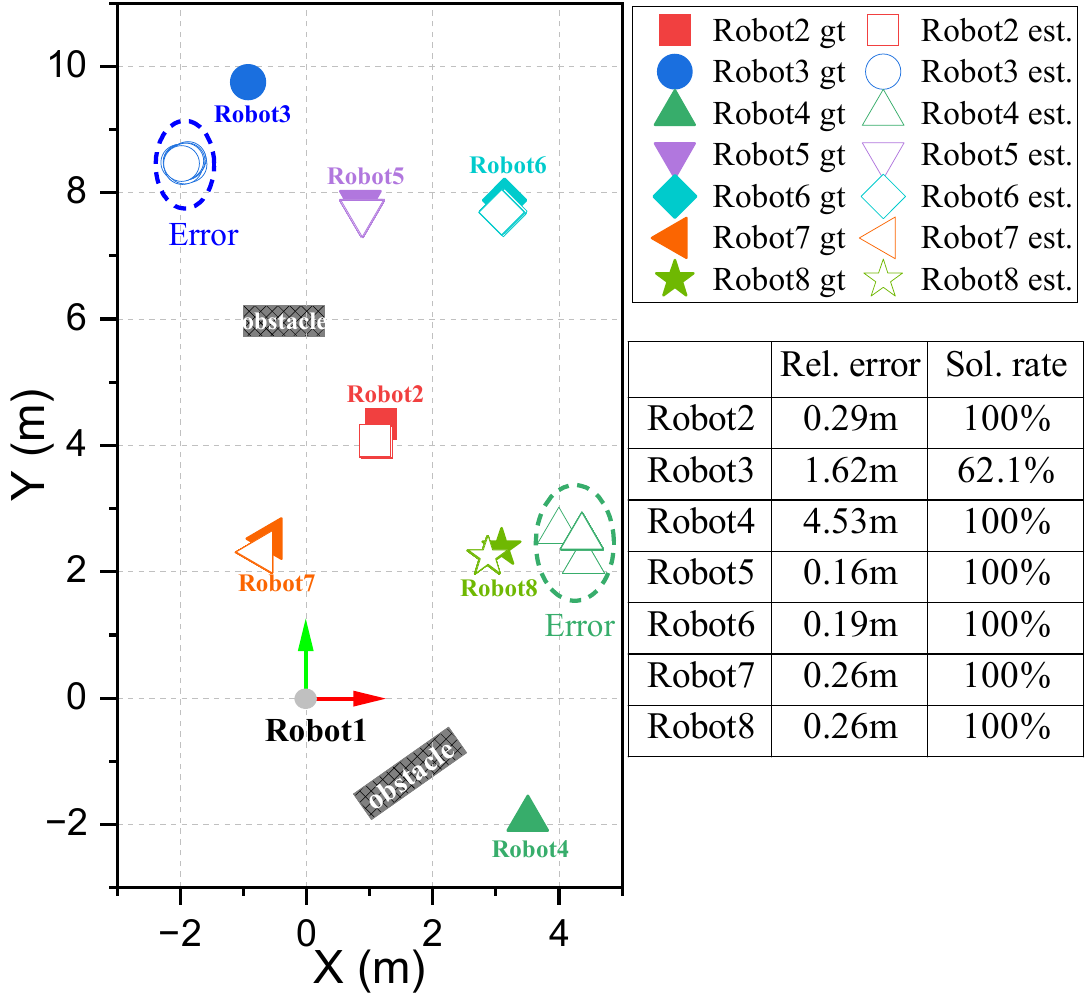}
    \label{distance-threshold-1.5}
}
\caption{The experimental results obtained from the optimal vertex correspondences disclosure based on common subgraph matching module (Section \ref{subsec-mcs-matching}) with different distance threshold $\lambda$ in Robot1's perspective. "-" denotes the robot was not estimated.}
\label{figure-distance-threshold}
\end{figure}

The relative position error and solution rate are summarized in Fig. \ref{figure-distance-threshold}.
For $\lambda = 0.2$m, Robot2 and Robot4 exhibit decreased relative position accuracy due to the false matches stemming from the optimal vertex correspondences disclosure based on common subgraph matching module.
In addition, although Robot8 achieved a localization accuracy of 0.26m, its solution rate was only 3.3\%, which means that \textit{Module 2} cannot provide a reliable solution for Robot8 during every trial.
For $\lambda = 0.5$m, the robots in LOS scenarios (i.e., Robot 5, Robot 6, Robot 7, and Robot 8) obtained good positioning accuracy and high solution rate, with around 0.25m and 100\%, respectively.
The occluded Robot4 was still falsely-matched and achieved a positioning accuracy of 6.43m, while another occluded Robot3 was not matched through \textit{Module 2}.
For $\lambda = 1.0$m, the performance of the optimal vertex correspondences disclosure based on common subgraph matching module improves when compared to $\lambda = 0.5$m. 
We observed that Robot2 is correctly matched throughout the experiment, and other robots in LOS (i.e., Robot5, Robot6, Robot7, and Robot8) are accurately estimated with a solution rate of 100\%.
When we increase $\lambda$ to 1.5m, the robots in LOS achieve a positioning accuracy of around 0.25m and a solution rate of 100\%.

A larger distance threshold $\lambda $ generally causes looser screening conditions, allowing more vertices to be considered in the matching process.
However, it produces some falsely-matched results with errors greater than 0.5 meters and an increase in computational cost.
With different distance threshold settings ($\lambda=0.2$, $\lambda=0.5$, $\lambda=1.0$, and $\lambda=1.5$), the average time consumed for executing the optimal vertex-corresponding matching module is 0.03s, 0.13s, 0.36s, and 2.84s, respectively.
Therefore, a too large or too small $\lambda $ obviously results in degradation in the performance of the optimal vertex correspondences disclosure based on common subgraph matching module.

\subsubsection{Falsely-matched robot recognition module (Module 3)}
\label{impact-similarity}
We examine the performance of the falsely-matched robot identification module described in Section \ref{subsec-mismatched-pair-identification}, which is used to eliminate the falsely-matched results produced by \textit{Module 2}.
Based on the conclusion presented in the previous subsection, we configured the distance threshold $\lambda$ to be 1.0m, and the communication rate was chosen to be 100\%.
With this configuration, the robots in LOS (Robot2, Robot5, Robot6, Robot7, and Robot8) were all accurately estimated by \textit{Module 2}, while the occluded Robot4 showed a poor performance.

\begin{table}[b]
\centering
\setlength{\abovecaptionskip}{0pt}%
\setlength{\belowcaptionskip}{5pt}%
\caption{Experimental evaluation of the proposed falsely-matched robot identification module (see Section \ref{subsec-mismatched-pair-identification}) at different dissimilarity threshold $\epsilon$. We aim to remove all falsely-matched results for Robot4 (a solution rate of 0\%) and retain the results for other robots (a solution rate of 100\%).}
\label{table-wrong-matched}
\setlength{\tabcolsep}{1.1mm}{
\begin{tabular}{|c|ccccccc|}
\hline
\multirow{2}{*}{} & \multicolumn{7}{c|}{\begin{tabular}[c]{@{}c@{}}The solution rate (\%) to localize nearby robots \\ at different dissimilarity threshold $\epsilon$ with respect to Robot1\end{tabular}}                                                                                                                  \\ \cline{2-8} 
                  & \multicolumn{1}{c|}{Robot2} & \multicolumn{1}{c|}{Robot3} & \multicolumn{1}{c|}{\textbf{Robot4}} & \multicolumn{1}{c|}{Robot5} & \multicolumn{1}{c|}{Robot6} & \multicolumn{1}{c|}{Robot7} & \textbf{Robot8} \\ \hline
$\epsilon=0.0$    & \multicolumn{1}{c|}{100\%}  & \multicolumn{1}{c|}{0\%}    & \multicolumn{1}{c|}{\textbf{0\%}}    & \multicolumn{1}{c|}{100\%}  & \multicolumn{1}{c|}{100\%}  & \multicolumn{1}{c|}{100\%}  & \textbf{97.7\%} \\ \hline
$\epsilon=0.2$    & \multicolumn{1}{c|}{100\%}  & \multicolumn{1}{c|}{0\%}    & \multicolumn{1}{c|}{\textbf{0\%}}    & \multicolumn{1}{c|}{100\%}  & \multicolumn{1}{c|}{100\%}  & \multicolumn{1}{c|}{100\%}  & \textbf{100\%}  \\ \hline
$\epsilon=0.5$    & \multicolumn{1}{c|}{100\%}  & \multicolumn{1}{c|}{0\%}    & \multicolumn{1}{c|}{\textbf{0\%}}    & \multicolumn{1}{c|}{100\%}  & \multicolumn{1}{c|}{100\%}  & \multicolumn{1}{c|}{100\%}  & \textbf{100\%}  \\ \hline
$\epsilon=0.7$    & \multicolumn{1}{c|}{100\%}  & \multicolumn{1}{c|}{0\%}    & \multicolumn{1}{c|}{\textbf{0\%}}    & \multicolumn{1}{c|}{100\%}  & \multicolumn{1}{c|}{100\%}  & \multicolumn{1}{c|}{100\%}  & \textbf{100\%}  \\ \hline
$\epsilon=1.0$    & \multicolumn{1}{c|}{100\%}  & \multicolumn{1}{c|}{0\%}    & \multicolumn{1}{c|}{\textbf{0\%}}    & \multicolumn{1}{c|}{100\%}  & \multicolumn{1}{c|}{100\%}  & \multicolumn{1}{c|}{100\%}  & \textbf{100\%}  \\ \hline
$\epsilon=1.3$    & \multicolumn{1}{c|}{100\%}  & \multicolumn{1}{c|}{0\%}    & \multicolumn{1}{c|}{\textbf{65.4\%}} & \multicolumn{1}{c|}{100\%}  & \multicolumn{1}{c|}{100\%}  & \multicolumn{1}{c|}{100\%}  & \textbf{100\%}  \\ \hline
$\epsilon=1.5$    & \multicolumn{1}{c|}{100\%}  & \multicolumn{1}{c|}{0\%}    & \multicolumn{1}{c|}{\textbf{98.5\%}} & \multicolumn{1}{c|}{100\%}  & \multicolumn{1}{c|}{100\%}  & \multicolumn{1}{c|}{100\%}  & \textbf{100\%}  \\ \hline
\end{tabular}}
\end{table}

The falsely-matched robot identification module aims to remove those falsely-matched robots (a solution rate of 0\%) and keep the well-matched robots (a solution rate of 100\%).
In this experiment, we would like to remove the results of both Robot3 and Robot4 and keep the results of other correctly matched robots.
Table \ref{table-wrong-matched} lists the solution rate of all robots after performing the falsely-matched robot identification module with different dissimilarity threshold $\epsilon$.
As we can see from this table, $\epsilon = 0.0$ gives a solution rate of 0\% for Robot4, meaning that all results of Robot4 are rejected.
However, the solution rate of Robot8 decreases from 100\% to 97.7\%, indicating that some accurate results are eliminated.
For $0.2\leq \epsilon \leq 1.0$, the falsely-matched robot identification module provides the best performance.
All erroneous results associated with the occluded Robot4 were accurately identified, whereas the correct results for Robot 2, Robot 5, Robot 6, Robot 7, and Robot 8 were preserved.
When $\epsilon $ is increased to 1.3 or more, the performance degrades, leading to the failure to identify all wrong results for Robot4.
Hence, we must select an appropriate value for the dissimilarity threshold $\epsilon $.
In general, while the distance threshold $\lambda$ and dissimilarity threshold $\epsilon $ can influence the performance of the approach, a well-chosen configuration is effective in common scenarios.
We are also exploring an approach with adaptive parameters in future work.

\subsubsection{Unestimated robots positioning module (Module 4)}
\label{re-localize unestimated robots}
\begin{figure*}
\centering
\setlength{\belowcaptionskip}{-5pt}
\subfloat[Module 2]{
    \includegraphics[width=55mm]{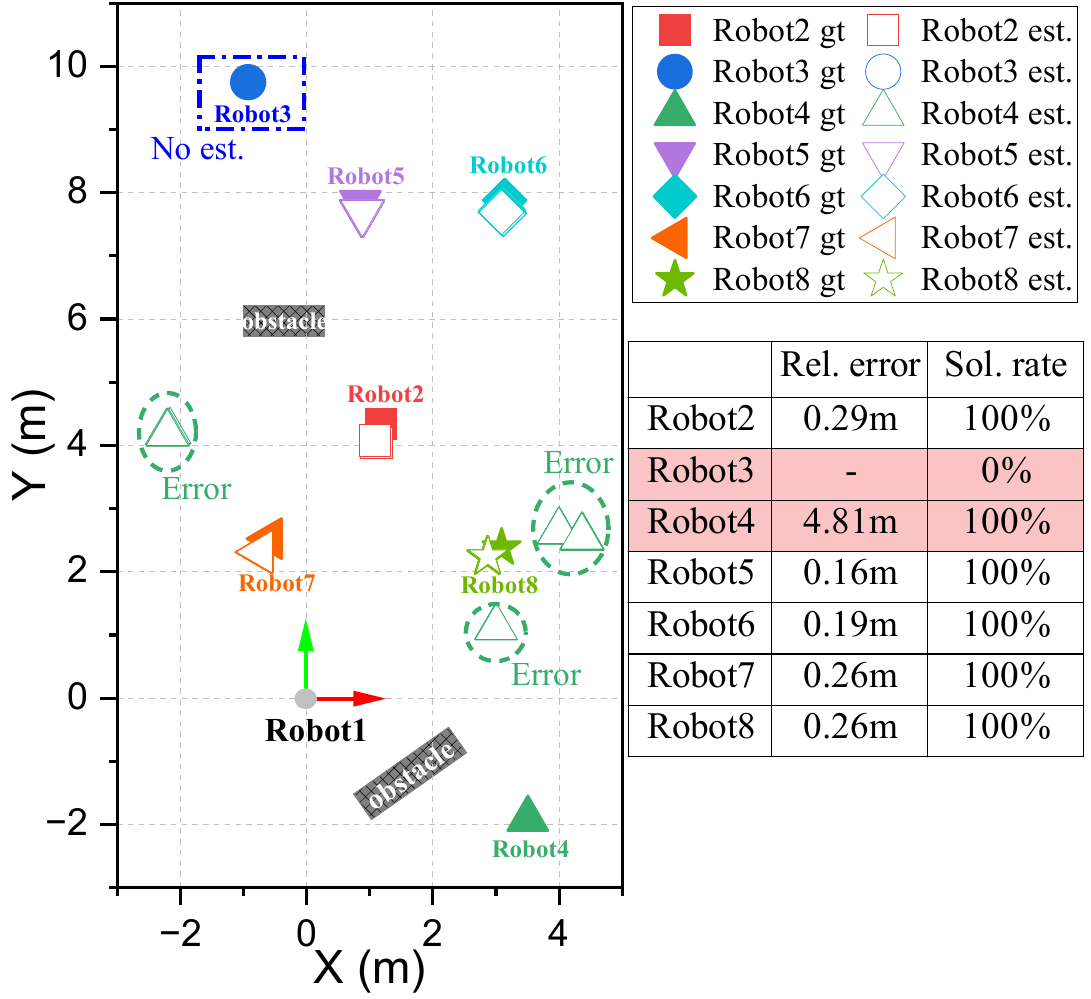}
    \label{figure-positioning-unestimated-robots-sub-1}
}
\subfloat[Module 2 + Module 3]{
    \includegraphics[width=55mm]{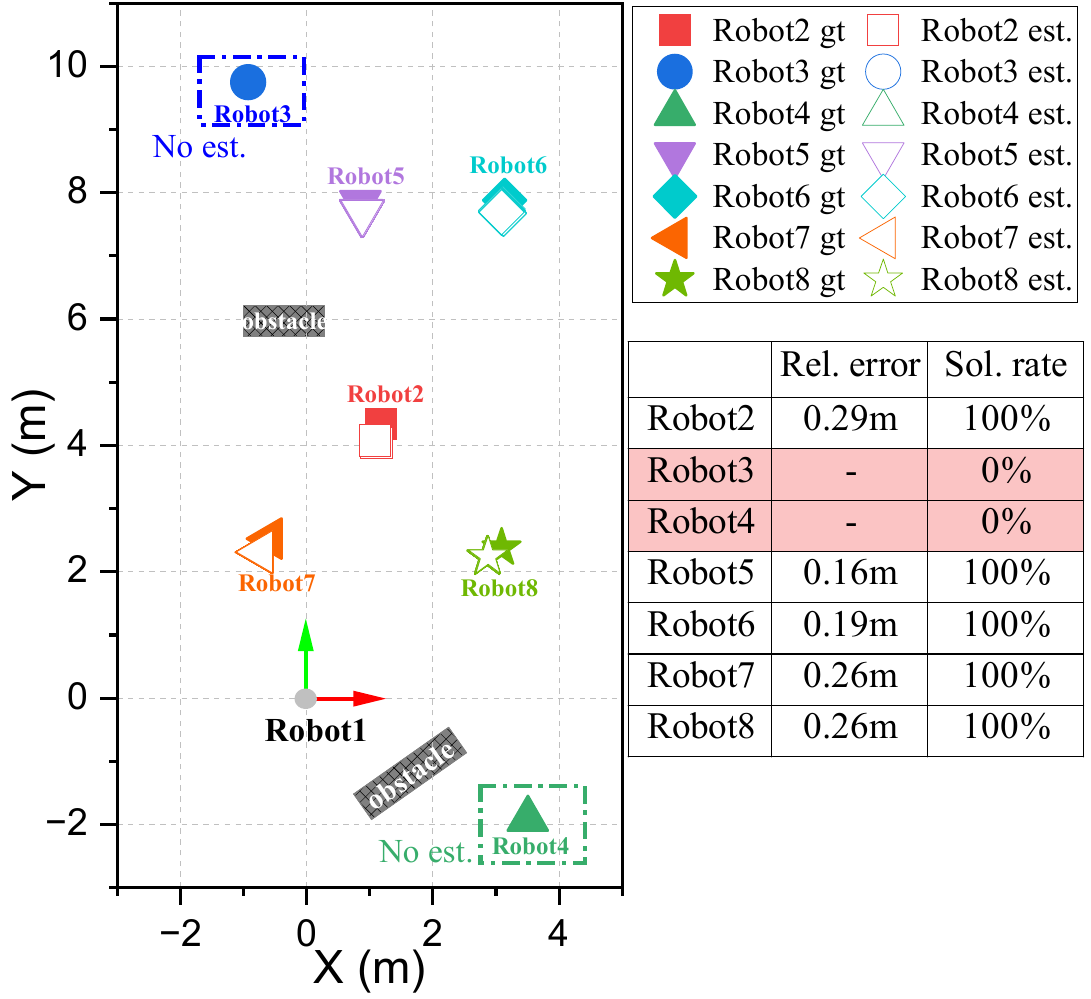}
    \label{figure-positioning-unestimated-robots-sub-2}
}
\subfloat[Module 2 + Module 3 + Module 4]{
    \includegraphics[width=55mm]{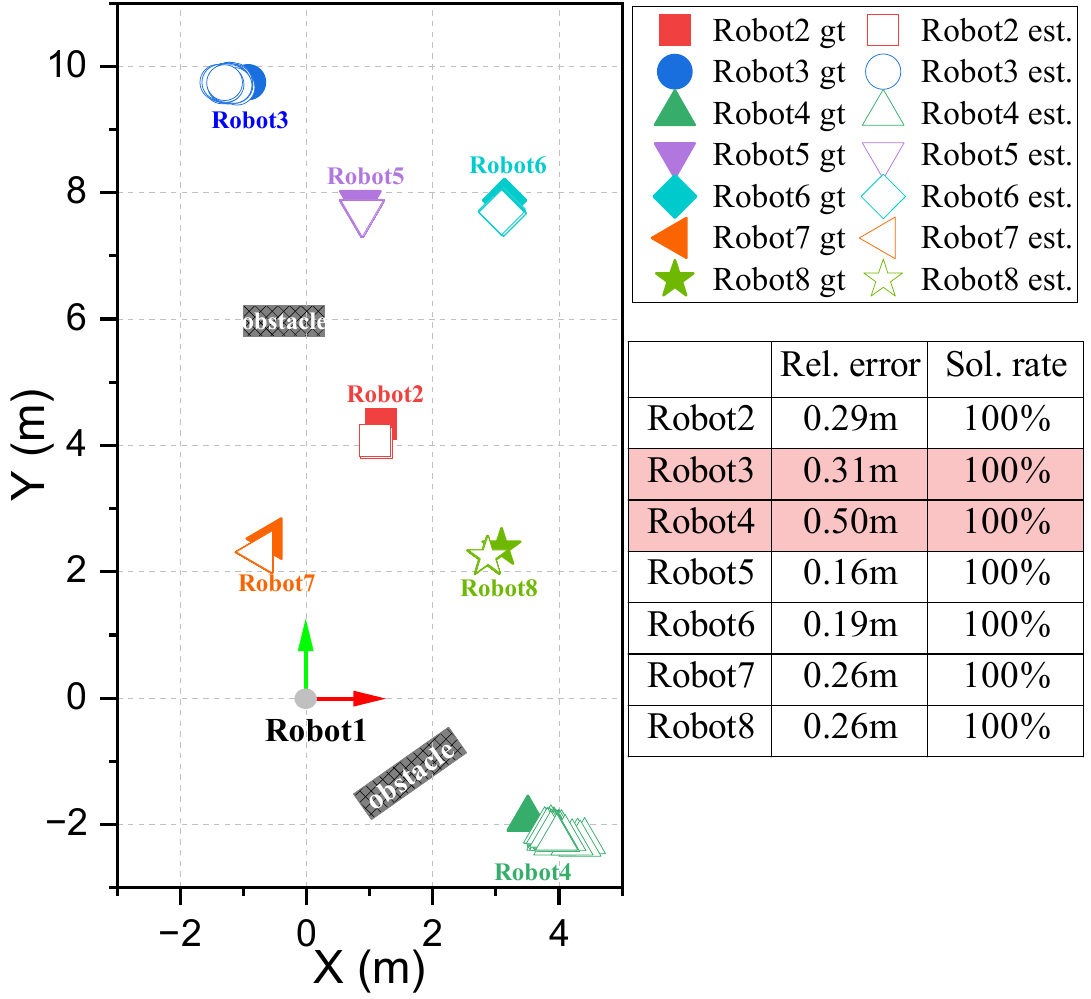}
    \label{figure-positioning-unestimated-robots-sub-3}
}
\caption{Visualization of experimental results: (a) results of the optimal vertex correspondences disclosure based on common subgraph matching module (Section \ref{subsec-mcs-matching}), (b) results with the falsely-matched robot identification module (Section \ref{subsec-mismatched-pair-identification}), (c) results with the positioning unestimated robots module (Section \ref{subsec-positioning-unmatched-robots}). "-" denotes the robot was not estimated.}
\label{figure-positioning-unestimated-robots}
\end{figure*}

Here, we present the evaluation of the unestimated robots positioning module demonstrated in Section \ref{subsec-positioning-unmatched-robots}, which is used to localize these robots that are not estimated through $\textit{Module 2}$ or are deemed suspicious results via $\textit{Module 3}$.
We set the distance threshold and dissimilarity threshold to 1.0 and 0.2, respectively.
The communication rate was still 100\%.
Fig. \ref{figure-positioning-unestimated-robots} visualizes the results obtained from \textit{Module 4} and its preparation phase (\textit{Module 2} and \textit{Module 3}).
As can be seen from Fig. \ref{figure-positioning-unestimated-robots-sub-2}, Robot3 and Robot4 are the unestimated robots.
The outcomes are shown in Fig. \ref{figure-positioning-unestimated-robots-sub-3}, both Robot3 and Robot4 were accurately localized based on the unestimated robots positioning module, achieving a solution rate of 100\% and positioning accuracy of approximately 0.5m.
It is worth noting that even though Robot3 and Robot4 are NLOS to Robot1, they may be LOS to other robots, and these UWB ranging measurements can thus offer valuable constraints for localization with Equation 5.

\subsection{Evaluation with different communication rates}
\label{impact communication}
The performance of the proposed approach under different communication rates is assessed in this subsection.
The experimental setup is the same as Section \ref{subsection-evaluation-different-modules}.
According to Section \ref{impact-distance} and Section \ref{impact-similarity}, the algorithm parameters are set to $\lambda=1.0$ and $\epsilon=0.2$.
In our communication configuration, we use 2 Bytes to represent a UWB ranging, with maximum range of 255.996m and resolution of 0.004m.
The amount of data transmission per robot depends on the number of robots $N$ and the communication rate, i.e., $2(N-1)\times (communication \ rate)$ Bytes.
According to the data transmission capacity of the Nooploop UWB sensor (100 Bytes/packet), at the communication rate of 100\%, the theoretical maximum number of robots that a multi-robot system can support is $N=51$. It is computed based on $2(N-1)\times (communication\ rate)\leq 100$.
In this experiment, we adopted 8 robots.
Table \ref{table-amount-data} lists the detailed communication cost per robot with different communication rates.
A higher communication rate causes an increase in the computational time for our approach, as shown in Table \ref{table-amount-data}.

\begin{table}[]
\setlength{\abovecaptionskip}{5pt}%
\setlength{\belowcaptionskip}{5pt}%
\caption{The communication cost per robot and the average time consumption of the proposed approach (Module 2 + Module 3 + Module 4) under different communication rates.}
\label{table-amount-data}
\centering
\begin{tabular}{|c|c|c|c|c|c|}
\hline
Comm. rate         & 0\%   & 14\%  & 43\%  & 71\%  & 100\% \\ \hline
Comm. cost (Bytes) & 0     & 2     & 6     & 10    & 14    \\ \hline
Comp. time (Seconds)   & 0.038 & 0.056 & 0.175 & 0.372 & 0.382 \\ \hline
\end{tabular}
\end{table}

Table \ref{table-real-experiment} lists the experimental results of our approach under different communication rates.
As can be seen from this table, the relative positioning accuracy improves with more UWB ranging measurements being shared.
Specifically, at a communication rate of 0\%, the optimal vertex correspondences disclosure based on common subgraph matching module depicted in Section \ref{subsec-mcs-matching} encounters difficulties in estimating Robot3, and most robots were wrongly localized.
The results for the falsely-matched robot identification module and the unestimated robots positioning module are poor due to the sparse edges in the UWB connection graph.

For the communication rate of 43\%, we observed from Table \ref{table-real-experiment} that the occluded Robot4 achieved an average relative positioning accuracy of 4.52m and a solution rate of 98.07\% through the optimal vertex correspondences disclosure based on common subgraph matching module depicted in Section \ref{subsec-mcs-matching}.
Although most of the false results for Robot4 could be removed through the falsely-matched robot identification module in Section \ref{subsec-mismatched-pair-identification}, 7.05\% of false results remained.
This is because there were not enough UWB ranging measurements in $\mathcal{G}_{uwb}$.
Eventually, Robot3 and Robot4 could be localized by the positioning unestimated robots module mentioned in Section \ref{subsec-positioning-unmatched-robots}, achieving an average relative positioning accuracy of 0.37m and 1.04m, respectively.

\begin{table*}[t]
\centering
\setlength{\abovecaptionskip}{0pt}%
\setlength{\belowcaptionskip}{5pt}%
\caption{Evaluation of the proposed approach under different communication rates: relative position error (in meters) and solution rate (in \%) of each robot with respect to robot1. "-" denotes the robot was not estimated.}
\label{table-real-experiment}
\resizebox{\textwidth}{!}{
\begin{tabular}{|c|c|cccccccccccccc|}
\hline
\multirow{3}{*}{Approach}                                                                                                                  & \multirow{3}{*}{\begin{tabular}[c]{@{}c@{}}Comm.\\ Rate\end{tabular}} & \multicolumn{14}{c|}{Average relative position error (in meters) with respect to Robot1 and solution rate (\%)}                                                                                                                                                                                                                                                                                                                                                                                                                                                                                                                                                                                                                                                                                                                                                                                                                                                                                                                                                           \\ \cline{3-16} 
                                                                                                                                           &                                                                       & \multicolumn{2}{c|}{Robot2}                                                                                                                          & \multicolumn{2}{c|}{Robot3}                                                                                                                          & \multicolumn{2}{c|}{Robot4}                                                                                                                          & \multicolumn{2}{c|}{Robot5}                                                                                                                          & \multicolumn{2}{c|}{Robot6}                                                                                                                          & \multicolumn{2}{c|}{Robot7}                                                                                                                          & \multicolumn{2}{c|}{Robot8}                                                                                                     \\ \cline{3-16} 
                                                                                                                                           &                                                                       & \multicolumn{1}{c|}{\begin{tabular}[c]{@{}c@{}}Rel.\\ error\end{tabular}} & \multicolumn{1}{c|}{\begin{tabular}[c]{@{}c@{}}Sol.\\ rate\end{tabular}} & \multicolumn{1}{c|}{\begin{tabular}[c]{@{}c@{}}Rel.\\ error\end{tabular}} & \multicolumn{1}{c|}{\begin{tabular}[c]{@{}c@{}}Sol.\\ rate\end{tabular}} & \multicolumn{1}{c|}{\begin{tabular}[c]{@{}c@{}}Rel.\\ error\end{tabular}} & \multicolumn{1}{c|}{\begin{tabular}[c]{@{}c@{}}Sol.\\ rate\end{tabular}} & \multicolumn{1}{c|}{\begin{tabular}[c]{@{}c@{}}Rel.\\ error\end{tabular}} & \multicolumn{1}{c|}{\begin{tabular}[c]{@{}c@{}}Sol.\\ rate\end{tabular}} & \multicolumn{1}{c|}{\begin{tabular}[c]{@{}c@{}}Rel.\\ error\end{tabular}} & \multicolumn{1}{c|}{\begin{tabular}[c]{@{}c@{}}Sol.\\ rate\end{tabular}} & \multicolumn{1}{c|}{\begin{tabular}[c]{@{}c@{}}Rel.\\ error\end{tabular}} & \multicolumn{1}{c|}{\begin{tabular}[c]{@{}c@{}}Sol.\\ rate\end{tabular}} & \multicolumn{1}{c|}{\begin{tabular}[c]{@{}c@{}}Rel.\\ error\end{tabular}} & \begin{tabular}[c]{@{}c@{}}Sol.\\ rate\end{tabular} \\ \hline
\begin{tabular}[c]{@{}c@{}}Optimal vertex correspondences\\ disclosure based on common subgraph \\ matching module (Module 2)\end{tabular} & \multirow{3}{*}{0\%}                                                  & \multicolumn{1}{c|}{3.33}                                                 & \multicolumn{1}{c|}{100\%}                                               & \multicolumn{1}{c|}{-}                                                    & \multicolumn{1}{c|}{0\%}                                                 & \multicolumn{1}{c|}{6.44}                                                 & \multicolumn{1}{c|}{97.18\%}                                             & \multicolumn{1}{c|}{0.65}                                                 & \multicolumn{1}{c|}{100\%}                                               & \multicolumn{1}{c|}{0.19}                                                 & \multicolumn{1}{c|}{100\%}                                               & \multicolumn{1}{c|}{0.26}                                                 & \multicolumn{1}{c|}{100\%}                                               & \multicolumn{1}{c|}{1.14}                                                 & 100\%                                               \\ \cline{1-1} \cline{3-16} 
\begin{tabular}[c]{@{}c@{}}+ Falsely-matched robot\\ recognition module (Module 3)\end{tabular}                                            &                                                                       & \multicolumn{1}{c|}{3.33}                                                 & \multicolumn{1}{c|}{100\%}                                               & \multicolumn{1}{c|}{-}                                                    & \multicolumn{1}{c|}{0\%}                                                 & \multicolumn{1}{c|}{6.44}                                                 & \multicolumn{1}{c|}{97.18\%}                                             & \multicolumn{1}{c|}{0.65}                                                 & \multicolumn{1}{c|}{100\%}                                               & \multicolumn{1}{c|}{0.19}                                                 & \multicolumn{1}{c|}{100\%}                                               & \multicolumn{1}{c|}{0.26}                                                 & \multicolumn{1}{c|}{100\%}                                               & \multicolumn{1}{c|}{1.14}                                                 & 100\%                                               \\ \cline{1-1} \cline{3-16} 
\begin{tabular}[c]{@{}c@{}}+ Unestimated robots\\ positioning module (Module 4)\end{tabular}                                               &                                                                       & \multicolumn{1}{c|}{3.33}                                                 & \multicolumn{1}{c|}{100\%}                                               & \multicolumn{1}{c|}{-}                                                    & \multicolumn{1}{c|}{0\%}                                                 & \multicolumn{1}{c|}{6.44}                                                 & \multicolumn{1}{c|}{97.18\%}                                             & \multicolumn{1}{c|}{0.65}                                                 & \multicolumn{1}{c|}{100\%}                                               & \multicolumn{1}{c|}{0.19}                                                 & \multicolumn{1}{c|}{100\%}                                               & \multicolumn{1}{c|}{0.26}                                                 & \multicolumn{1}{c|}{100\%}                                               & \multicolumn{1}{c|}{1.14}                                                 & 100\%                                               \\ \hline
\begin{tabular}[c]{@{}c@{}}Optimal vertex correspondences\\ disclosure based on common subgraph \\ matching module (Module 2)\end{tabular} & \multirow{3}{*}{14\%}                                                 & \multicolumn{1}{c|}{3.31}                                                 & \multicolumn{1}{c|}{100\%}                                               & \multicolumn{1}{c|}{-}                                                    & \multicolumn{1}{c|}{0\%}                                                 & \multicolumn{1}{c|}{6.42}                                                 & \multicolumn{1}{c|}{98.32\%}                                             & \multicolumn{1}{c|}{0.16}                                                 & \multicolumn{1}{c|}{100\%}                                               & \multicolumn{1}{c|}{0.19}                                                 & \multicolumn{1}{c|}{100\%}                                               & \multicolumn{1}{c|}{0.26}                                                 & \multicolumn{1}{c|}{100\%}                                               & \multicolumn{1}{c|}{0.26}                                                 & 100\%                                               \\ \cline{1-1} \cline{3-16} 
\begin{tabular}[c]{@{}c@{}}+ Falsely-matched robot\\ recognition module (Module 3)\end{tabular}                                            &                                                                       & \multicolumn{1}{c|}{3.31}                                                 & \multicolumn{1}{c|}{100\%}                                               & \multicolumn{1}{c|}{-}                                                    & \multicolumn{1}{c|}{0\%}                                                 & \multicolumn{1}{c|}{6.42}                                                 & \multicolumn{1}{c|}{98.32\%}                                             & \multicolumn{1}{c|}{0.16}                                                 & \multicolumn{1}{c|}{100\%}                                               & \multicolumn{1}{c|}{0.19}                                                 & \multicolumn{1}{c|}{100\%}                                               & \multicolumn{1}{c|}{0.26}                                                 & \multicolumn{1}{c|}{100\%}                                               & \multicolumn{1}{c|}{0.26}                                                 & 100\%                                               \\ \cline{1-1} \cline{3-16} 
\begin{tabular}[c]{@{}c@{}}+ Unestimated robots\\ positioning module (Module 4)\end{tabular}                                               &                                                                       & \multicolumn{1}{c|}{3.31}                                                 & \multicolumn{1}{c|}{100\%}                                               & \multicolumn{1}{c|}{20.58}                                                & \multicolumn{1}{c|}{98.32\%}                                             & \multicolumn{1}{c|}{6.42}                                                 & \multicolumn{1}{c|}{98.32\%}                                             & \multicolumn{1}{c|}{0.16}                                                 & \multicolumn{1}{c|}{100\%}                                               & \multicolumn{1}{c|}{0.19}                                                 & \multicolumn{1}{c|}{100\%}                                               & \multicolumn{1}{c|}{0.26}                                                 & \multicolumn{1}{c|}{100\%}                                               & \multicolumn{1}{c|}{0.26}                                                 & 100\%                                               \\ \hline
\begin{tabular}[c]{@{}c@{}}Optimal vertex correspondences\\ disclosure based on common subgraph \\ matching module (Module 2)\end{tabular} & \multirow{3}{*}{43\%}                                                 & \multicolumn{1}{c|}{0.29}                                                 & \multicolumn{1}{c|}{100\%}                                               & \multicolumn{1}{c|}{\textbf{5.48}}                                        & \multicolumn{1}{c|}{\textbf{0.64\%}}                                     & \multicolumn{1}{c|}{\textbf{4.52}}                                        & \multicolumn{1}{c|}{\textbf{98.07\%}}                                    & \multicolumn{1}{c|}{0.15}                                                 & \multicolumn{1}{c|}{100\%}                                               & \multicolumn{1}{c|}{0.19}                                                 & \multicolumn{1}{c|}{100\%}                                               & \multicolumn{1}{c|}{0.26}                                                 & \multicolumn{1}{c|}{100\%}                                               & \multicolumn{1}{c|}{0.26}                                                 & 100\%                                               \\ \cline{1-1} \cline{3-16} 
\begin{tabular}[c]{@{}c@{}}+ Falsely-matched robot\\ recognition module (Module 3)\end{tabular}                                            &                                                                       & \multicolumn{1}{c|}{0.29}                                                 & \multicolumn{1}{c|}{100\%}                                               & \multicolumn{1}{c|}{\textbf{-}}                                           & \multicolumn{1}{c|}{\textbf{0\%}}                                        & \multicolumn{1}{c|}{\textbf{3.72}}                                        & \multicolumn{1}{c|}{\textbf{7.05\%}}                                     & \multicolumn{1}{c|}{0.15}                                                 & \multicolumn{1}{c|}{100\%}                                               & \multicolumn{1}{c|}{0.19}                                                 & \multicolumn{1}{c|}{100\%}                                               & \multicolumn{1}{c|}{0.26}                                                 & \multicolumn{1}{c|}{100\%}                                               & \multicolumn{1}{c|}{0.26}                                                 & 100\%                                               \\ \cline{1-1} \cline{3-16} 
\begin{tabular}[c]{@{}c@{}}+ Unestimated robots\\ positioning module (Module 4)\end{tabular}                                               &                                                                       & \multicolumn{1}{c|}{0.29}                                                 & \multicolumn{1}{c|}{100\%}                                               & \multicolumn{1}{c|}{\textbf{0.37}}                                        & \multicolumn{1}{c|}{\textbf{100\%}}                                      & \multicolumn{1}{c|}{\textbf{1.04}}                                        & \multicolumn{1}{c|}{\textbf{100\%}}                                      & \multicolumn{1}{c|}{0.15}                                                 & \multicolumn{1}{c|}{100\%}                                               & \multicolumn{1}{c|}{0.19}                                                 & \multicolumn{1}{c|}{100\%}                                               & \multicolumn{1}{c|}{0.26}                                                 & \multicolumn{1}{c|}{100\%}                                               & \multicolumn{1}{c|}{0.26}                                                 & 100\%                                               \\ \hline
\begin{tabular}[c]{@{}c@{}}Optimal vertex correspondences\\ disclosure based on common subgraph \\ matching module (Module 2)\end{tabular} & \multirow{3}{*}{71\%}                                                 & \multicolumn{1}{c|}{0.29}                                                 & \multicolumn{1}{c|}{100\%}                                               & \multicolumn{1}{c|}{-}                                                    & \multicolumn{1}{c|}{0\%}                                                 & \multicolumn{1}{c|}{\textbf{4.64}}                                        & \multicolumn{1}{c|}{\textbf{100\%}}                                      & \multicolumn{1}{c|}{0.16}                                                 & \multicolumn{1}{c|}{100\%}                                               & \multicolumn{1}{c|}{0.19}                                                 & \multicolumn{1}{c|}{100\%}                                               & \multicolumn{1}{c|}{0.26}                                                 & \multicolumn{1}{c|}{100\%}                                               & \multicolumn{1}{c|}{0.26}                                                 & 100\%                                               \\ \cline{1-1} \cline{3-16} 
\begin{tabular}[c]{@{}c@{}}+ Falsely-matched robot\\ recognition module (Module 3)\end{tabular}                                            &                                                                       & \multicolumn{1}{c|}{0.29}                                                 & \multicolumn{1}{c|}{100\%}                                               & \multicolumn{1}{c|}{-}                                                    & \multicolumn{1}{c|}{0\%}                                                 & \multicolumn{1}{c|}{\textbf{-}}                                           & \multicolumn{1}{c|}{\textbf{0\%}}                                        & \multicolumn{1}{c|}{0.16}                                                 & \multicolumn{1}{c|}{100\%}                                               & \multicolumn{1}{c|}{0.19}                                                 & \multicolumn{1}{c|}{100\%}                                               & \multicolumn{1}{c|}{0.26}                                                 & \multicolumn{1}{c|}{100\%}                                               & \multicolumn{1}{c|}{0.26}                                                 & 100\%                                               \\ \cline{1-1} \cline{3-16} 
\begin{tabular}[c]{@{}c@{}}+ Unestimated robots\\ positioning module (Module 4)\end{tabular}                                               &                                                                       & \multicolumn{1}{c|}{\textbf{0.29}}                                        & \multicolumn{1}{c|}{\textbf{100\%}}                                      & \multicolumn{1}{c|}{\textbf{0.31}}                                        & \multicolumn{1}{c|}{\textbf{100\%}}                                      & \multicolumn{1}{c|}{\textbf{0.53}}                                        & \multicolumn{1}{c|}{\textbf{100\%}}                                      & \multicolumn{1}{c|}{\textbf{0.16}}                                        & \multicolumn{1}{c|}{\textbf{100\%}}                                      & \multicolumn{1}{c|}{\textbf{0.19}}                                        & \multicolumn{1}{c|}{\textbf{100\%}}                                      & \multicolumn{1}{c|}{\textbf{0.26}}                                        & \multicolumn{1}{c|}{\textbf{100\%}}                                      & \multicolumn{1}{c|}{\textbf{0.26}}                                        & \textbf{100\%}                                      \\ \hline
\begin{tabular}[c]{@{}c@{}}Optimal vertex correspondences\\ disclosure based on common subgraph \\ matching module (Module 2)\end{tabular} & \multirow{3}{*}{100\%}                                                & \multicolumn{1}{c|}{0.29}                                                 & \multicolumn{1}{c|}{100\%}                                               & \multicolumn{1}{c|}{-}                                                    & \multicolumn{1}{c|}{0\%}                                                 & \multicolumn{1}{c|}{4.54}                                                 & \multicolumn{1}{c|}{100\%}                                               & \multicolumn{1}{c|}{0.16}                                                 & \multicolumn{1}{c|}{100\%}                                               & \multicolumn{1}{c|}{0.19}                                                 & \multicolumn{1}{c|}{100\%}                                               & \multicolumn{1}{c|}{0.26}                                                 & \multicolumn{1}{c|}{100\%}                                               & \multicolumn{1}{c|}{0.26}                                                 & 100\%                                               \\ \cline{1-1} \cline{3-16} 
\begin{tabular}[c]{@{}c@{}}+ Falsely-matched robot\\ recognition module (Module 3)\end{tabular}                                            &                                                                       & \multicolumn{1}{c|}{0.29}                                                 & \multicolumn{1}{c|}{100\%}                                               & \multicolumn{1}{c|}{-}                                                    & \multicolumn{1}{c|}{0\%}                                                 & \multicolumn{1}{c|}{-}                                                    & \multicolumn{1}{c|}{0\%}                                                 & \multicolumn{1}{c|}{0.16}                                                 & \multicolumn{1}{c|}{100\%}                                               & \multicolumn{1}{c|}{0.19}                                                 & \multicolumn{1}{c|}{100\%}                                               & \multicolumn{1}{c|}{0.26}                                                 & \multicolumn{1}{c|}{100\%}                                               & \multicolumn{1}{c|}{0.26}                                                 & 100\%                                               \\ \cline{1-1} \cline{3-16} 
\begin{tabular}[c]{@{}c@{}}+ Unestimated robots\\ positioning module (Module 4)\end{tabular}                                               &                                                                       & \multicolumn{1}{c|}{0.29}                                                 & \multicolumn{1}{c|}{100\%}                                               & \multicolumn{1}{c|}{0.31}                                                 & \multicolumn{1}{c|}{100\%}                                               & \multicolumn{1}{c|}{0.50}                                                 & \multicolumn{1}{c|}{100\%}                                               & \multicolumn{1}{c|}{0.16}                                                 & \multicolumn{1}{c|}{100\%}                                               & \multicolumn{1}{c|}{0.19}                                                 & \multicolumn{1}{c|}{100\%}                                               & \multicolumn{1}{c|}{0.26}                                                 & \multicolumn{1}{c|}{100\%}                                               & \multicolumn{1}{c|}{0.26}                                                 & 100\%                                               \\ \hline
\end{tabular}}
\end{table*}

When the communication rate raised to 71\%, the performance reaches its peak, exhibiting high reliability and accuracy.
As can be seen from Table \ref{table-real-experiment}, the robots in LOS (i.e., Robot2, Robot5, Robot6, Robot7, and Robot8) achieve a relative positioning accuracy of under 0.3m, while those in NLOS obtain a relative positioning accuracy of approximately 0.5m.
This is because both the well-matched robots and the UWB ranging measurements are noisy.
Moreover, the solution rate for each robot is increased to 100\%.
Compared to the communication rate of 43\%, Robot4's positioning accuracy is improved by 50\%, from 1.04m to 0.53m.
This positioning accuracy may not be suitable for some high-precision tasks, but it remains effective for applications like search and rescue or multi-robot exploration.
Additionally, there is no significant improvement in the performance from the communication rate of 71\% to 100\%.
This is mainly because the UWB ranging measurements are already enough to construct $\mathcal{G}_{uwb}$ for localization when the communication rate is 71\%.

\subsection{Evaluation under different UWB communication protocols}
\label{subsection-different-sorting-methods}
In this subsection, we discuss the results of the proposed approach using different UWB communication protocols, such as fixed order and random order, for sharing UWB ranging measurements.
Fig. \ref{figure-communication-protocol} illustrates the fixed order, while the random order indicates that each robot randomly selects a portion of UWB ranging measurements from all available measurements to share, based on varying communication rates.
As observed in Table \ref{table-order-experiment}, the performance of both fixed order and random order grows with the communication rate increases.
When the communication rate is below 43\%, both protocols show poor results.
Once the communication rate reaches or exceeds 43\%, the fixed order consistently outperforms the random order and achieves stable results.
This occurs because the random order may reduce the connectivity and density of the UWB connection graph under a limited communication rate, as robots redundantly share UWB ranging with the same attributes (e.g., $r_{2,3}$ and $r_{3,2}$).
Thus, to ensure robustness and accuracy, a fixed order, as depicted in Fig. \ref{figure-communication-protocol}, is adopted for each robot to share UWB ranging measurements.

\begin{table*}[t]
\centering
\setlength{\abovecaptionskip}{0pt}%
\setlength{\belowcaptionskip}{0pt}%
\caption{Evaluation of the proposed approach (Module 2 + Module 3 + Module 4) under different UWB communication protocols: relative position error (in meters) and solution rate (in \%) of each robot with respect to robot1. "-" denotes the robot was not estimated.}
\label{table-order-experiment}
\resizebox{\textwidth}{!}{
\begin{tabular}{|c|c|cccccccccccccc|}
\hline
\multirow{3}{*}{\begin{tabular}[c]{@{}c@{}}Protocols\end{tabular}} & \multirow{3}{*}{\begin{tabular}[c]{@{}c@{}}Comm.\\ Rate\end{tabular}} & \multicolumn{14}{c|}{Average relative position error (in meters) with respect to Robot1 and solution rate (\%)}                                                                                                                                                                                                                                                                                                                                                                                                                                                                                                                                                                                                                                                                                                                                                                                                                                                                                                                                                           \\ \cline{3-16} 
                                                                        &                                                                       & \multicolumn{2}{c|}{Robot2}                                                                                                                          & \multicolumn{2}{c|}{Robot3}                                                                                                                          & \multicolumn{2}{c|}{Robot4}                                                                                                                          & \multicolumn{2}{c|}{Robot5}                                                                                                                          & \multicolumn{2}{c|}{Robot6}                                                                                                                          & \multicolumn{2}{c|}{Robot7}                                                                                                                          & \multicolumn{2}{c|}{Robot8}                                                                                                     \\ \cline{3-16} 
                                                                        &                                                                       & \multicolumn{1}{c|}{\begin{tabular}[c]{@{}c@{}}Rel.\\ error\end{tabular}} & \multicolumn{1}{c|}{\begin{tabular}[c]{@{}c@{}}Sol.\\ rate\end{tabular}} & \multicolumn{1}{c|}{\begin{tabular}[c]{@{}c@{}}Rel.\\ error\end{tabular}} & \multicolumn{1}{c|}{\begin{tabular}[c]{@{}c@{}}Sol.\\ rate\end{tabular}} & \multicolumn{1}{c|}{\begin{tabular}[c]{@{}c@{}}Rel.\\ error\end{tabular}} & \multicolumn{1}{c|}{\begin{tabular}[c]{@{}c@{}}Sol.\\ rate\end{tabular}} & \multicolumn{1}{c|}{\begin{tabular}[c]{@{}c@{}}Rel.\\ error\end{tabular}} & \multicolumn{1}{c|}{\begin{tabular}[c]{@{}c@{}}Sol.\\ rate\end{tabular}} & \multicolumn{1}{c|}{\begin{tabular}[c]{@{}c@{}}Rel.\\ error\end{tabular}} & \multicolumn{1}{c|}{\begin{tabular}[c]{@{}c@{}}Sol.\\ rate\end{tabular}} & \multicolumn{1}{c|}{\begin{tabular}[c]{@{}c@{}}Rel.\\ error\end{tabular}} & \multicolumn{1}{c|}{\begin{tabular}[c]{@{}c@{}}Sol.\\ rate\end{tabular}} & \multicolumn{1}{c|}{\begin{tabular}[c]{@{}c@{}}Rel.\\ error\end{tabular}} & \begin{tabular}[c]{@{}c@{}}Sol.\\ rate\end{tabular} \\ \hline
Fixed                                                                   & \multirow{2}{*}{0\%}                                                  & \multicolumn{1}{c|}{3.33}                                                 & \multicolumn{1}{c|}{100\%}                                               & \multicolumn{1}{c|}{-}                                                    & \multicolumn{1}{c|}{0\%}                                                 & \multicolumn{1}{c|}{6.44}                                                 & \multicolumn{1}{c|}{97.18\%}                                             & \multicolumn{1}{c|}{0.65}                                                 & \multicolumn{1}{c|}{100\%}                                               & \multicolumn{1}{c|}{0.19}                                                 & \multicolumn{1}{c|}{100\%}                                               & \multicolumn{1}{c|}{0.26}                                                 & \multicolumn{1}{c|}{100\%}                                               & \multicolumn{1}{c|}{1.14}                                                 & 100\%                                               \\ \cline{1-1} \cline{3-16} 
Random                                                                  &                                                                       & \multicolumn{1}{c|}{3.33}                                                 & \multicolumn{1}{c|}{100\%}                                               & \multicolumn{1}{c|}{-}                                                    & \multicolumn{1}{c|}{0\%}                                                 & \multicolumn{1}{c|}{6.44}                                                 & \multicolumn{1}{c|}{97.18\%}                                             & \multicolumn{1}{c|}{0.65}                                                 & \multicolumn{1}{c|}{100\%}                                               & \multicolumn{1}{c|}{0.19}                                                 & \multicolumn{1}{c|}{100\%}                                               & \multicolumn{1}{c|}{0.26}                                                 & \multicolumn{1}{c|}{100\%}                                               & \multicolumn{1}{c|}{1.14}                                                 & 100\%                                               \\ \hline
Fixed                                                          & \multirow{2}{*}{14\%}                                                 & \multicolumn{1}{c|}{\textbf{3.31}}                                        & \multicolumn{1}{c|}{\textbf{100\%}}                                      & \multicolumn{1}{c|}{\textbf{20.58}}                                       & \multicolumn{1}{c|}{\textbf{98.32\%}}                                    & \multicolumn{1}{c|}{\textbf{6.42}}                                        & \multicolumn{1}{c|}{\textbf{98.32\%}}                                    & \multicolumn{1}{c|}{\textbf{0.16}}                                        & \multicolumn{1}{c|}{\textbf{100\%}}                                      & \multicolumn{1}{c|}{\textbf{0.19}}                                        & \multicolumn{1}{c|}{\textbf{100\%}}                                      & \multicolumn{1}{c|}{\textbf{0.26}}                                        & \multicolumn{1}{c|}{\textbf{100\%}}                                      & \multicolumn{1}{c|}{\textbf{0.26}}                                        & \textbf{100\%}                                      \\ \cline{1-1} \cline{3-16} 
Random                                                                  &                                                                       & \multicolumn{1}{c|}{1.41}                                                 & \multicolumn{1}{c|}{99.34\%}                                             & \multicolumn{1}{c|}{9.97}                                                 & \multicolumn{1}{c|}{38.41\%}                                             & \multicolumn{1}{c|}{4.86}                                                 & \multicolumn{1}{c|}{76.82\%}                                             & \multicolumn{1}{c|}{0.29}                                                 & \multicolumn{1}{c|}{99.34\%}                                             & \multicolumn{1}{c|}{0.32}                                                 & \multicolumn{1}{c|}{100\%}                                               & \multicolumn{1}{c|}{0.72}                                                 & \multicolumn{1}{c|}{100\%}                                               & \multicolumn{1}{c|}{0.56}                                                 & 99.34\%                                             \\ \hline
Fixed                                                          & \multirow{2}{*}{43\%}                                                 & \multicolumn{1}{c|}{\textbf{0.29}}                                        & \multicolumn{1}{c|}{\textbf{100\%}}                                      & \multicolumn{1}{c|}{\textbf{0.37}}                                        & \multicolumn{1}{c|}{\textbf{100\%}}                                      & \multicolumn{1}{c|}{\textbf{1.04}}                                        & \multicolumn{1}{c|}{\textbf{100\%}}                                      & \multicolumn{1}{c|}{\textbf{0.15}}                                        & \multicolumn{1}{c|}{\textbf{100\%}}                                      & \multicolumn{1}{c|}{\textbf{0.19}}                                        & \multicolumn{1}{c|}{\textbf{100\%}}                                      & \multicolumn{1}{c|}{\textbf{0.26}}                                        & \multicolumn{1}{c|}{\textbf{100\%}}                                      & \multicolumn{1}{c|}{\textbf{0.26}}                                        & \textbf{100\%}                                      \\ \cline{1-1} \cline{3-16} 
Random                                                                  &                                                                       & \multicolumn{1}{c|}{0.67}                                                 & \multicolumn{1}{c|}{99.31\%}                                             & \multicolumn{1}{c|}{1.79}                                                 & \multicolumn{1}{c|}{100\%}                                               & \multicolumn{1}{c|}{1.62}                                                 & \multicolumn{1}{c|}{97.93\%}                                             & \multicolumn{1}{c|}{0.56}                                                 & \multicolumn{1}{c|}{100\%}                                               & \multicolumn{1}{c|}{0.64}                                                 & \multicolumn{1}{c|}{100\%}                                               & \multicolumn{1}{c|}{0.48}                                                 & \multicolumn{1}{c|}{100\%}                                               & \multicolumn{1}{c|}{0.52}                                                 & 97.93\%                                             \\ \hline
Fixed                                                          & \multirow{2}{*}{71\%}                                                 & \multicolumn{1}{c|}{\textbf{0.29}}                                        & \multicolumn{1}{c|}{\textbf{100\%}}                                      & \multicolumn{1}{c|}{\textbf{0.31}}                                        & \multicolumn{1}{c|}{\textbf{100\%}}                                      & \multicolumn{1}{c|}{\textbf{0.53}}                                        & \multicolumn{1}{c|}{\textbf{100\%}}                                      & \multicolumn{1}{c|}{\textbf{0.16}}                                        & \multicolumn{1}{c|}{\textbf{100\%}}                                      & \multicolumn{1}{c|}{\textbf{0.19}}                                        & \multicolumn{1}{c|}{\textbf{100\%}}                                      & \multicolumn{1}{c|}{\textbf{0.26}}                                        & \multicolumn{1}{c|}{\textbf{100\%}}                                      & \multicolumn{1}{c|}{\textbf{0.26}}                                        & \textbf{100\%}                                      \\ \cline{1-1} \cline{3-16} 
Random                                                                  &                                                                       & \multicolumn{1}{c|}{0.42}                                                 & \multicolumn{1}{c|}{100\%}                                               & \multicolumn{1}{c|}{0.63}                                                 & \multicolumn{1}{c|}{100\%}                                               & \multicolumn{1}{c|}{0.68}                                                 & \multicolumn{1}{c|}{98.58\%}                                             & \multicolumn{1}{c|}{0.41}                                                 & \multicolumn{1}{c|}{100\%}                                               & \multicolumn{1}{c|}{0.42}                                                 & \multicolumn{1}{c|}{100\%}                                               & \multicolumn{1}{c|}{0.36}                                                 & \multicolumn{1}{c|}{100\%}                                               & \multicolumn{1}{c|}{0.36}                                                 & 100\%                                               \\ \hline
Fixed                                                                   & \multirow{2}{*}{100\%}                                                & \multicolumn{1}{c|}{0.29}                                                 & \multicolumn{1}{c|}{100\%}                                               & \multicolumn{1}{c|}{0.31}                                                 & \multicolumn{1}{c|}{100\%}                                               & \multicolumn{1}{c|}{0.50}                                                 & \multicolumn{1}{c|}{100\%}                                               & \multicolumn{1}{c|}{0.16}                                                 & \multicolumn{1}{c|}{100\%}                                               & \multicolumn{1}{c|}{0.19}                                                 & \multicolumn{1}{c|}{100\%}                                               & \multicolumn{1}{c|}{0.26}                                                 & \multicolumn{1}{c|}{100\%}                                               & \multicolumn{1}{c|}{0.26}                                                 & 100\%                                               \\ \cline{1-1} \cline{3-16} 
Random                                                                  &                                                                       & \multicolumn{1}{c|}{0.29}                                                 & \multicolumn{1}{c|}{100\%}                                               & \multicolumn{1}{c|}{0.31}                                                 & \multicolumn{1}{c|}{100\%}                                               & \multicolumn{1}{c|}{0.50}                                                 & \multicolumn{1}{c|}{100\%}                                               & \multicolumn{1}{c|}{0.16}                                                 & \multicolumn{1}{c|}{100\%}                                               & \multicolumn{1}{c|}{0.19}                                                 & \multicolumn{1}{c|}{100\%}                                               & \multicolumn{1}{c|}{0.26}                                                 & \multicolumn{1}{c|}{100\%}                                               & \multicolumn{1}{c|}{0.26}                                                 & 100\%                                               \\ \hline
\end{tabular}}
\end{table*}

\begin{figure}[h]
\centering
\vspace{-0.5cm}
\subfloat[$N$ = 6]{
    \includegraphics[width=40mm]{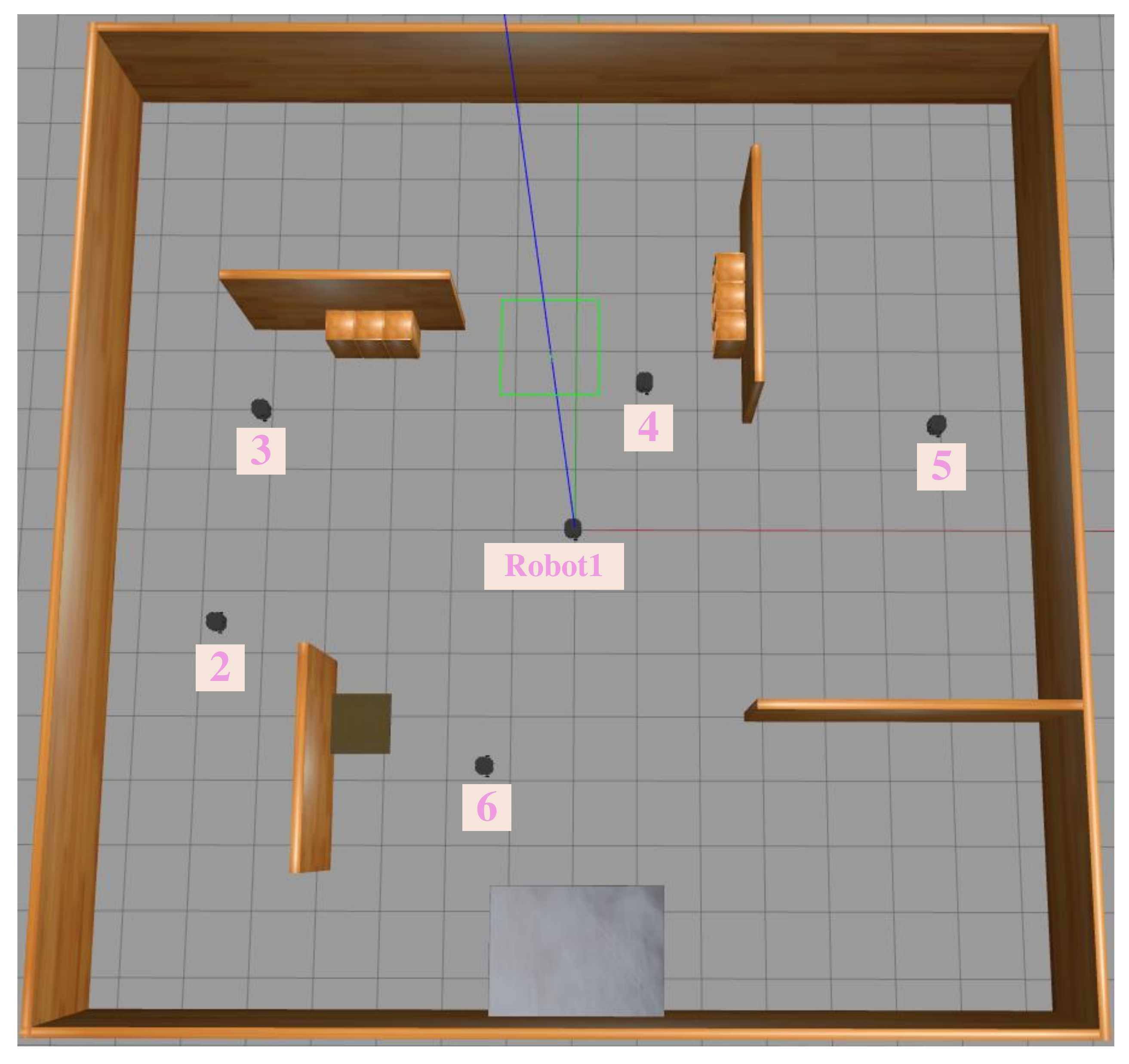}
    \label{6-robots-simulation}
}
\subfloat[$N$ = 8]{
    \includegraphics[width=40mm]{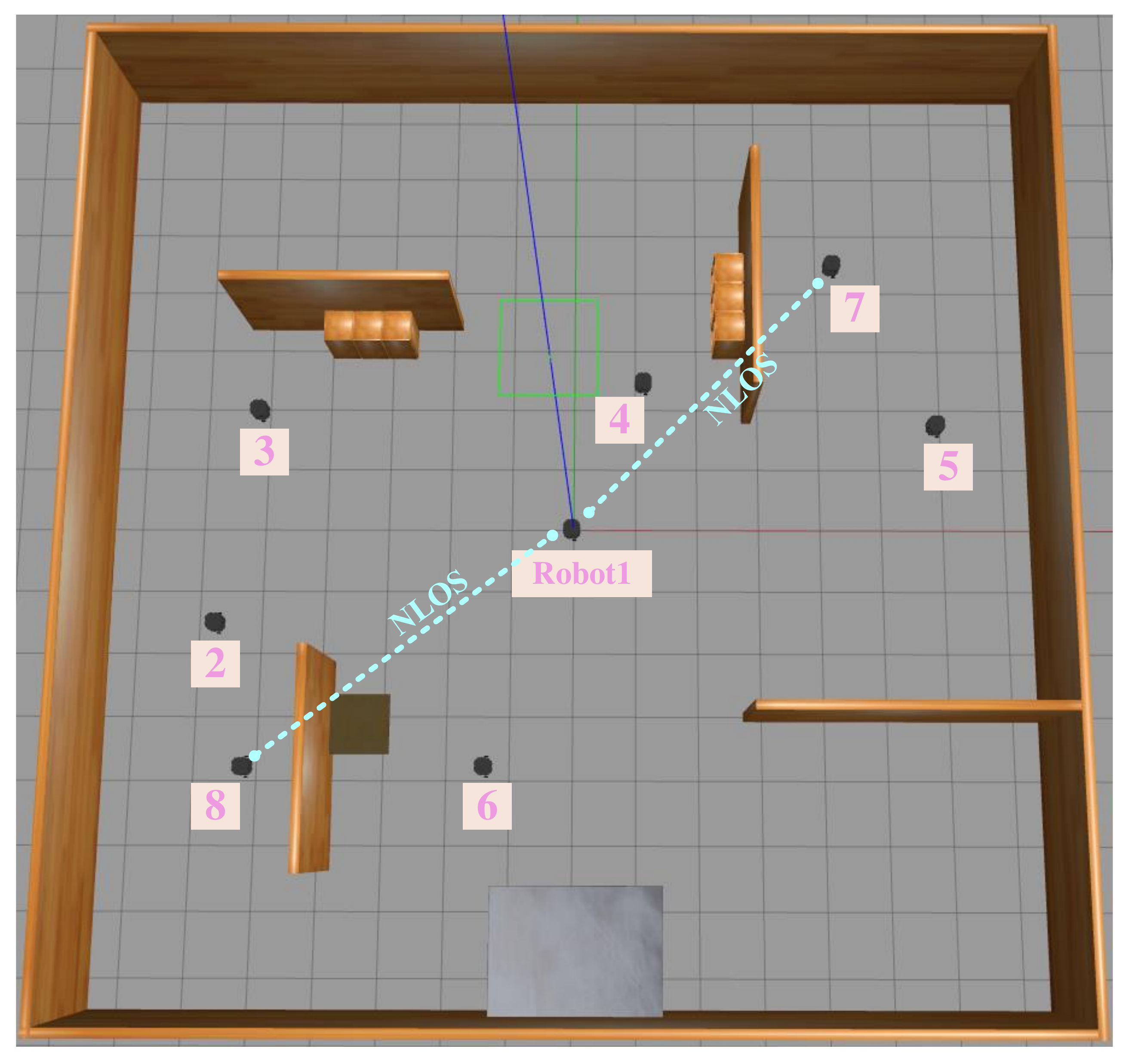}
    \label{8-robots-simulation}
}
\hfil
\subfloat[$N$ = 10]{
    \includegraphics[width=40mm]{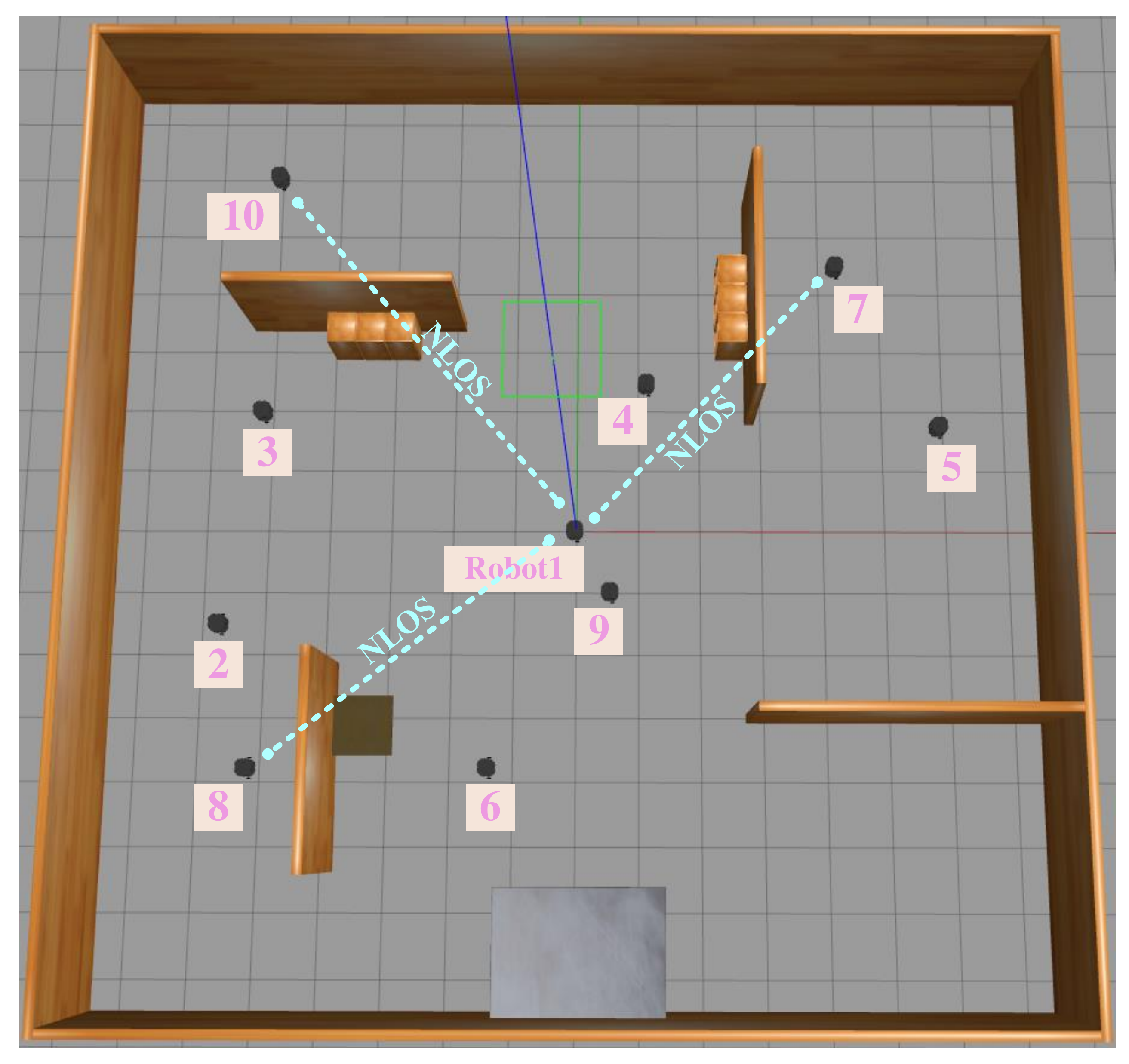}
    \label{10-robots-simulation}
}
\subfloat[$N$ = 12]{
    \includegraphics[width=40mm]{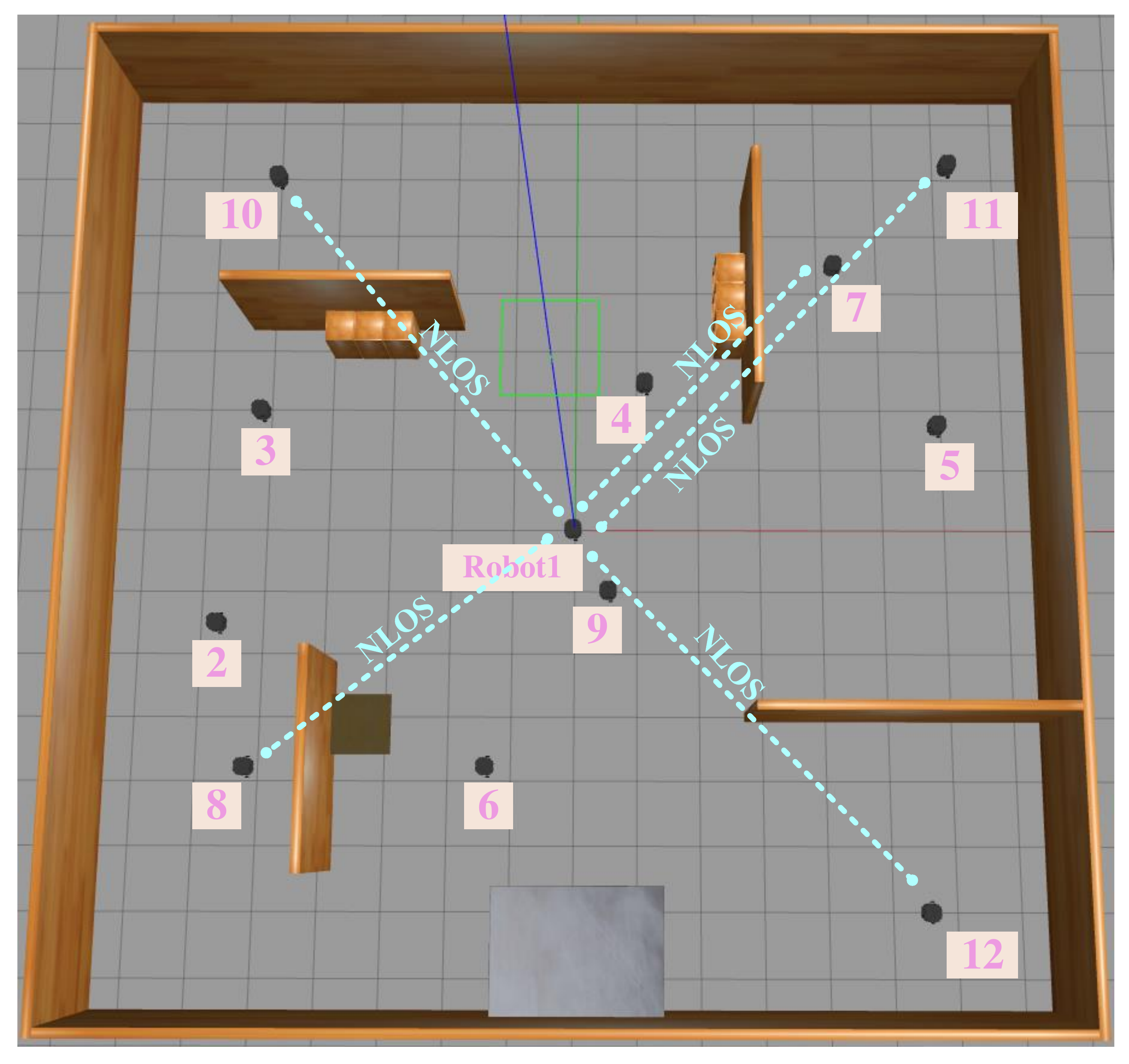}
    \label{12-robots-simulation}
}
\caption{The simulation setups in a 15m$\times$15m environment with different numbers of robots. }
\label{figure-robots-scale-environment}
\end{figure}

\vspace{-6pt}
\subsection{Evaluation with different numbers of robots}
\label{robots scale evaluation}
\begin{figure}[t]
\centering
\vspace{-0.5cm}
\subfloat[$N$ = 6]{
    \includegraphics[width=43mm]{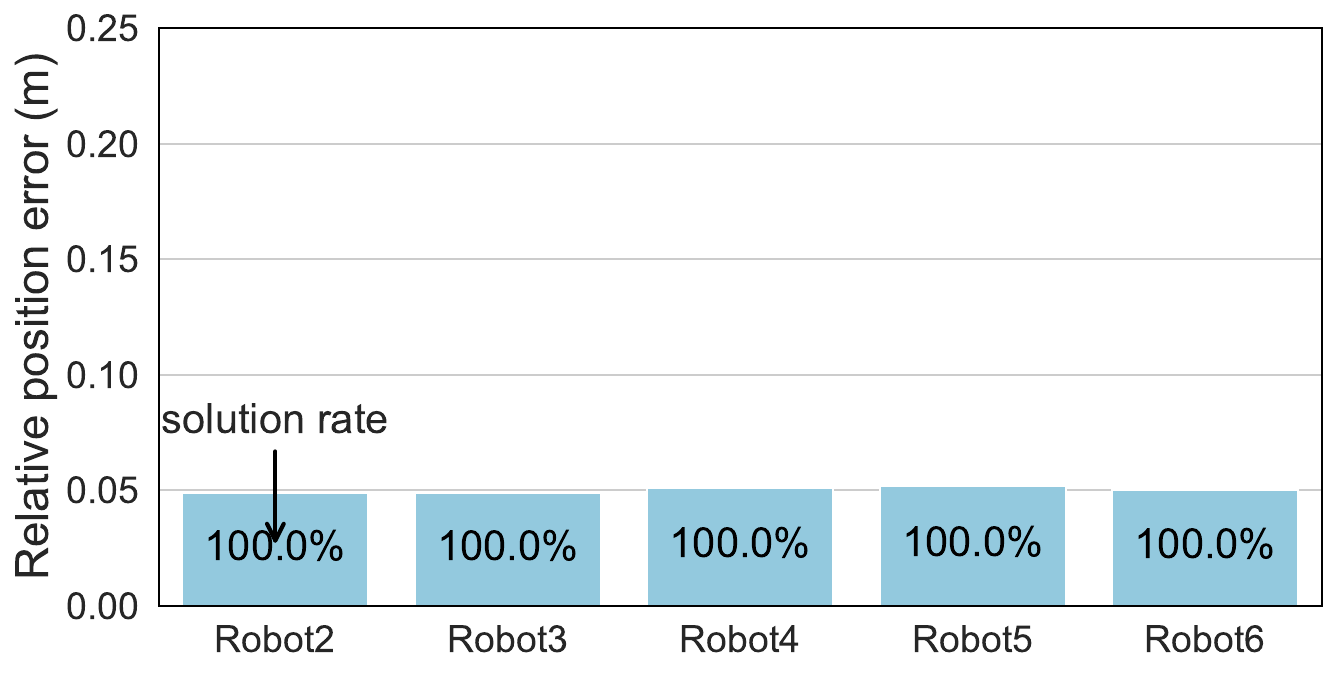}
    \label{6-robots-result}
}
\subfloat[$N$ = 8]{
    \includegraphics[width=43mm]{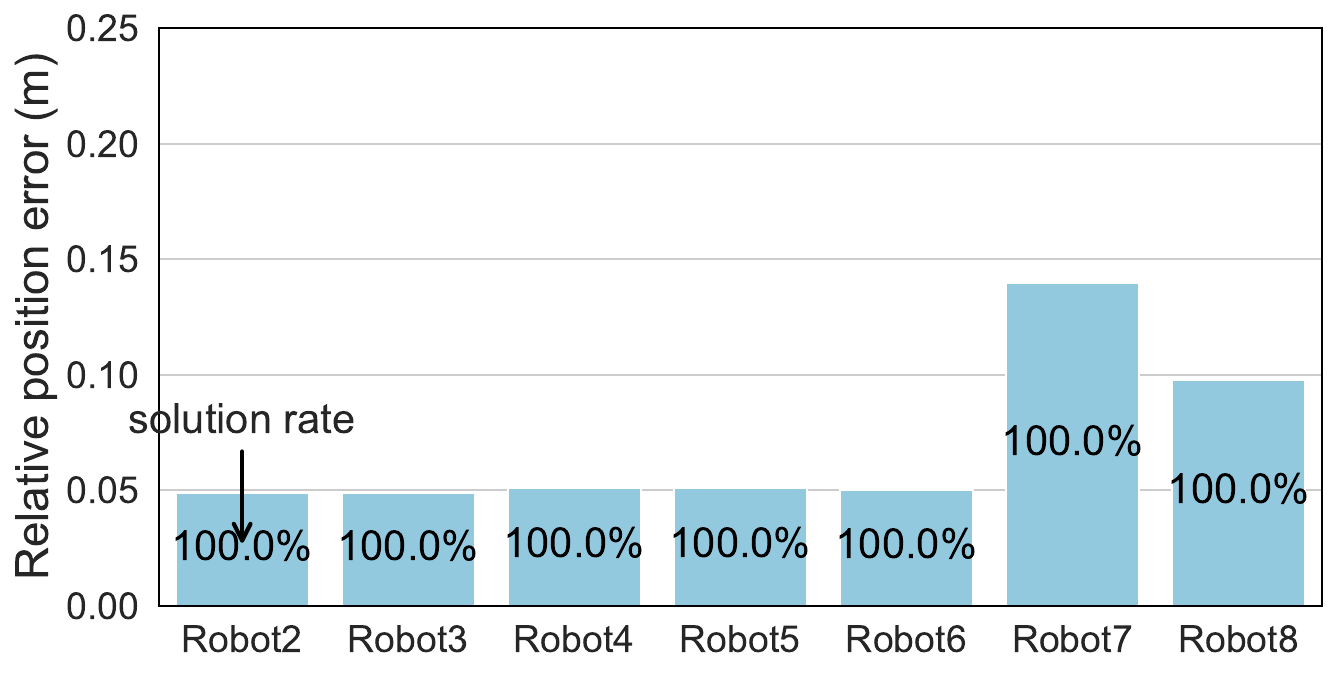}
    \label{8-robots-result}
}
\hfil
\subfloat[$N$ = 10]{
    \includegraphics[width=43mm]{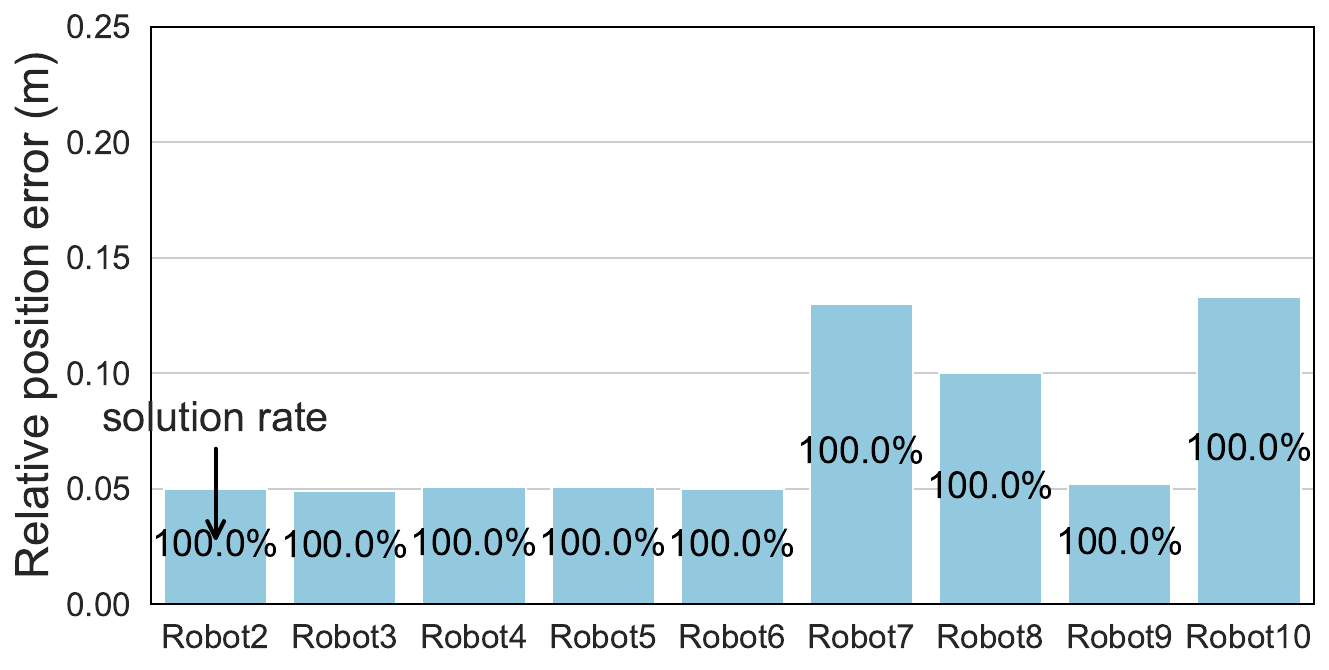}
    \label{10-robots-result}
}
\subfloat[$N$ = 12]{
    \includegraphics[width=43mm]{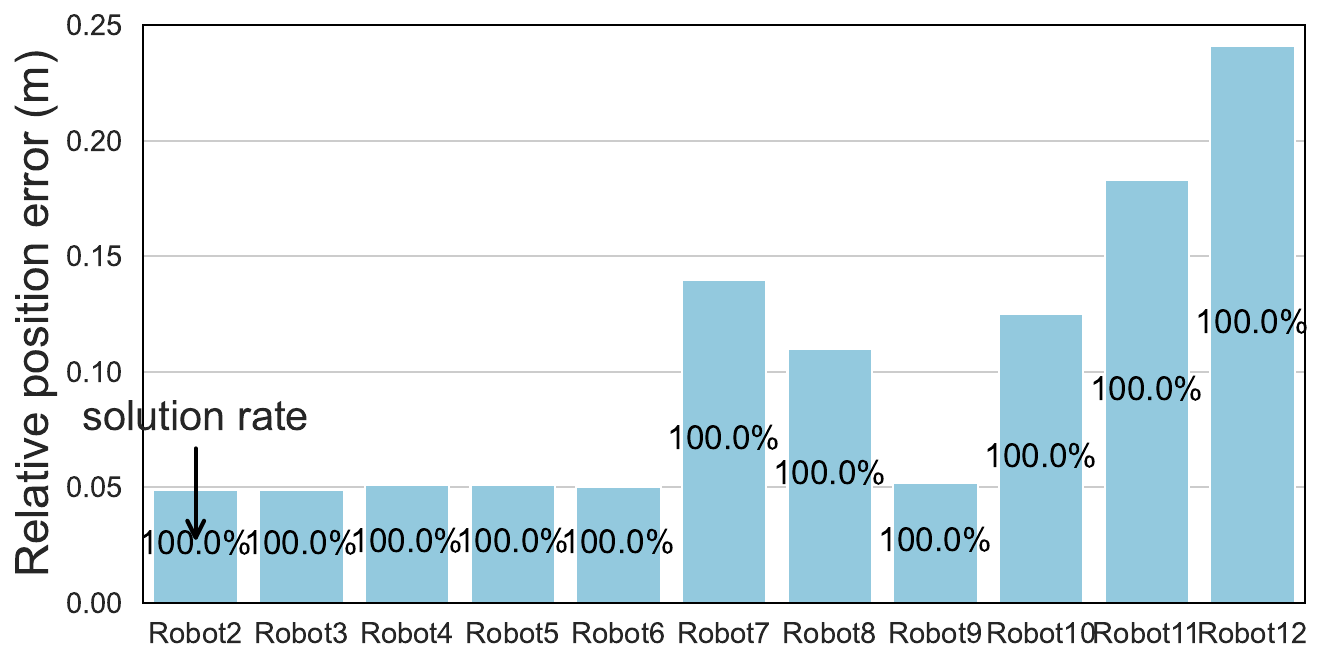}
    \label{12-robots-result}
}
\caption{Evaluation of the positioning accuracy of each robot relative to Robot1, with different numbers of robots.}
\label{figure-robots-scale-evaluation}
\end{figure}

This subsection presents the performance of the proposed approach with varying numbers of robots.
Due to limitations in experimental resources, we generated simulated data for 6-12 robots, as shown in Fig. \ref{figure-robots-scale-environment}.
Various obstacles, such as tables, cabinets, and walls, were included in the simulation environment.
The simulated LiDAR data had a maximum range of 10m, and UWB ranging measurements were modeled with Gaussian noise, $\mathcal{N}(0,\,0.3^{2})$.
The distance threshold $\lambda$ was set to 1.0m.
The dissimilarity threshold $\epsilon$ and communication rate were selected to be 0.2 and 71\%, respectively.
Fig. \ref{figure-robots-scale-evaluation} illustrates the estimation results using the proposed approach (\textit{Module 2 + Module 3 + Module 4}) for each robot relative to Robot1, with different numbers of robots.
The results demonstrate that our approach is able to accurately and effectively localize surrounding robots in varying $N$.
In LOS scenarios, we achieved an average relative position error of approximately 0.05m.
For robots in NLOS scenarios, the average relative position error is below 0.25m.
In addition, the proposed approach exhibits robust performance, consistently achieving a solution rate of 100\% across varying numbers of robots.
While the performance in terms of relative position accuracy and solution rate demonstrates robustness, the communication cost and computational time increases as the number of robots scales, as summarized in Table \ref{table-simulation-comm-comp}.

\begin{table}[t]
\setlength{\abovecaptionskip}{0pt}%
\setlength{\belowcaptionskip}{0pt}%
\caption{The communication cost per robot and the average time consumption of the proposed approach with different numbers of robots.}
\label{table-simulation-comm-comp}
\centering
\begin{tabular}{|c|c|c|c|c|}
\hline
$N$         & 6   & 8  & 10  & 12 \\ \hline
Comm. cost (Bytes) & 7     & 10     & 13     & 16 \\ \hline
Comp. time (Seconds)   & 0.107 & 0.207 & 1.03 & 1.725 \\ \hline
\end{tabular}
\end{table}

\subsection{Evaluation in dynamic scenario}
\label{evaluation dynamic scenarios}
\begin{figure}[t]
\centering
\includegraphics[width=70mm]{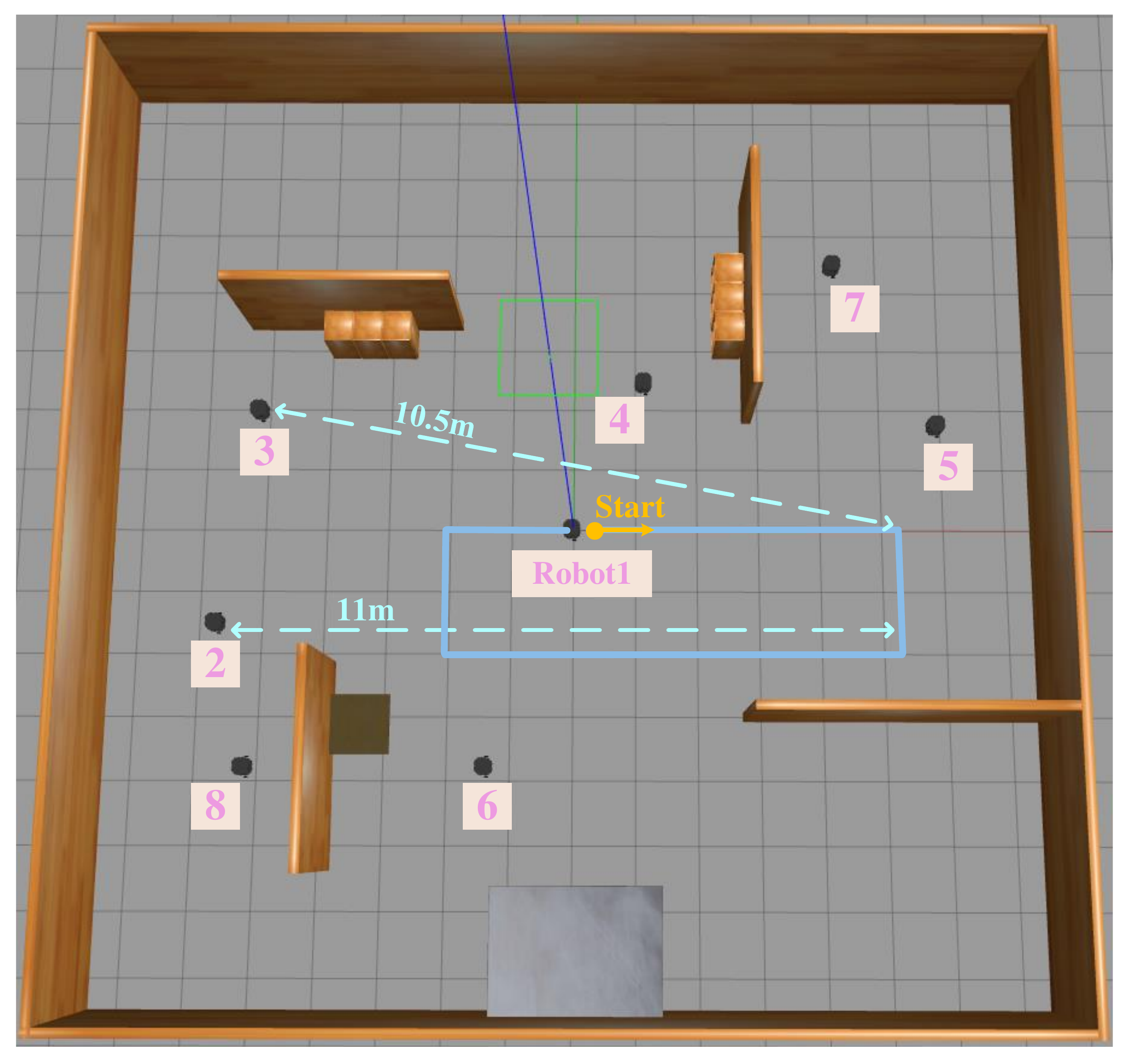}
\caption{Robot1 is controlled to move along the rectangle path. When Robot1 moves to the rightmost position, both Robot2 and Robot3 are out of Robot1's LiDAR range.}
\label{figure-moving-robots-environment}
\end{figure}

\begin{figure}[t]
\centering
\includegraphics[width=90mm]{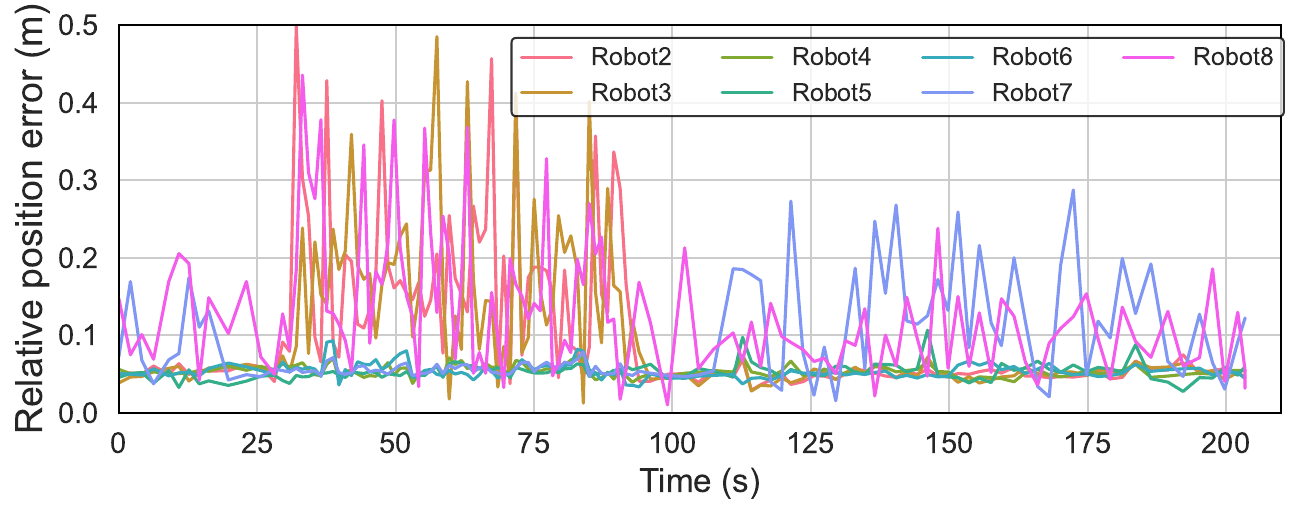}
\caption{The relative position error of each robot relative to Robot1 when Robot1 moves along the rectangle path.}
\label{figure-moving-robots-results}
\end{figure}

In this experiment, the proposed approach is evaluated in a dynamic scenario.
Specifically, Robot1 was controlled to move along the rectangle path with an average speed of 0.1m/s, as depicted in Fig. \ref{figure-moving-robots-environment}.
Notably, when Robot1 moved to the rightmost point, both Robot2 and Robot3 exceeded Robot1's LiDAR range (i.e., maximum range of 10m).
During the experiment, Robot1 transitioned from NLOS state to LOS state, and then reverted to NLOS.
To ensure a fair and effective evaluation, we set the algorithm parameters to $\lambda=1.0$, $\epsilon=0.2$, and the communication rate of 71\%.
Fig. \ref{figure-moving-robots-results} illustrates the relative position error of each robot with respect to Robot1 during the movement.
In the initial period, the occluded Robot7 and Robot8 were localized with an error of approximately 0.1m, while other robots in LOS maintained a stable and low error of 0.05m.
Between 30s and 90s, as Robot2 and Robot3 moved out of Robot1's LiDAR range and Robot7 transitioned to LOS, the former showed high positioning error, whereas the latter exhibited a low error.
After 110s, when Robot2 and Robot3 re-entered Robot1's LiDAR view and Robot7 was once again obstructed by a wall, Robot2 and Robot3 had a low positioning error of 0.05m, while Robot7's positioning error increased to 0.15m.
In summary, this experiment demonstrates the capacity of our approach to perform effectively in dynamic scenarios.

\section{Conclusion and Future works} \label{sec-conclusion}
This paper presented an approach for distributed relative position estimation using UWB ranging and LiDAR measurements without any external communication infrastructure.
We construct UWB connection graph and LiDAR connection graph using UWB ranging measurements and LiDAR cluster information, respectively.
We formulated a common subgraph matching solution for relative localization.
In order to ensure accuracy, we utilize the falsely-matched robot identification module to remove the suspicious results.
Finally, the localization of the unestimated robots is attained by leveraging the well-matched robots and the corresponding UWB ranging measurements.
We conducted a series of experiments in real and simulation environments to demonstrate the robustness and feasibility of the proposed approach.
The results show that the proposed approach is able to provide accurate relative positions for robots in both LOS and NLOS cases in communication-limited environments.
Moreover, our approach demonstrates the ability in scalability.
In future works, we plan to extend our approach to 3D applications using 3D LiDAR or camera, which will also facilitate efficient robot detection without requiring changes to the robot's structure.
Additionally, we aim to enhance the system's accuracy and robustness by improving UWB ranging stability \cite{UWB-stability} and strengthening spatial relationships among objects.

\begin{IEEEbiography}[{\includegraphics[width=1in,height=1.25in,clip,keepaspectratio]{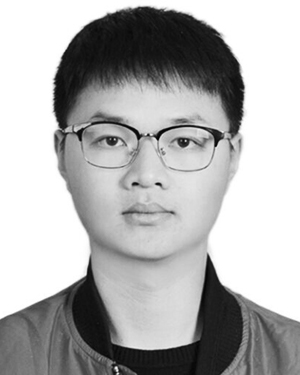}}]{Zhiqiang Cao}
received the B.S. and MA.Sc. degrees from the Southwest University of Science and Technology, Mianyang, China, in 2019 and 2022, respectively. He is currently pursuing his Ph.D. degree at the Singapore University of Technology and Design, Singapore, under the supervision of Prof. Chau Yuen. His current research interests include multi-robot systems, collaborative localization, and distributed SLAM systems.
\end{IEEEbiography}

\vspace{-1.8cm}

\begin{IEEEbiography}[{\includegraphics[width=1in,height=1.25in,clip,keepaspectratio]{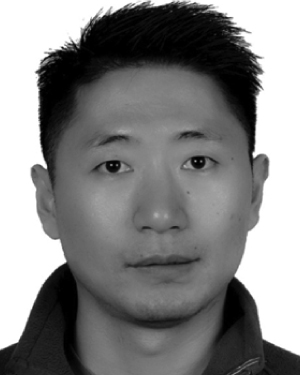}}]{Ran Liu}
(Senior Member, IEEE) received the B.S. degree from the Southwest University of Science and Technology, China, in 2007, and the Ph.D. degree from the University of Tüebingen, Germany, in 2014. From 2015 to 2024, he was a research fellow at the Singapore University of Technology and Design. He is currently a senior research fellow at the Nanyang Technological University. His research interests include robotics and SLAM.
\end{IEEEbiography}

\vspace{-1.8cm}

\begin{IEEEbiography}[{\includegraphics[width=1in,height=1.25in,clip,keepaspectratio]{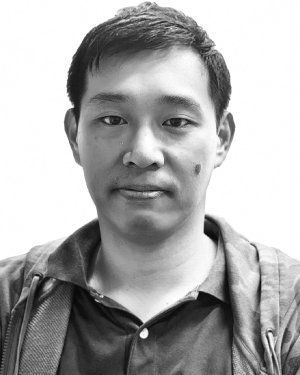}}]{Billy Pik Lik Lau}
(Senior Member, IEEE) received Ph.D. at the Singapore University of Technology and Design (SUTD) in 2021 as well as a degree in computer science and M. Phil degree in computer science from Curtin University in 2010 and 2014, respectively. He is currently a research fellow at SUTD Temasek lab. His Ph.D. research work includes Smart city, the Internet of things, data fusion, urban science, and big data analysis. His current interest is the integration of data fusion, urban science, multi-agent system design, multi-robotic collaboration, and robotic exploration.
\end{IEEEbiography}

\vspace{-1.8cm}

\begin{IEEEbiography}[{\includegraphics[width=1in,height=1.25in,clip,keepaspectratio]{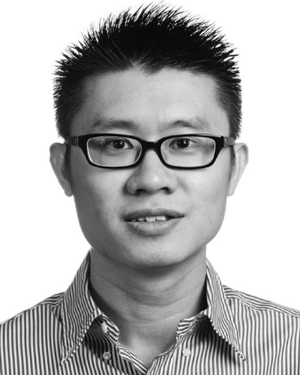}}]{Chau Yuen}
(Fellow, IEEE) received the B.Eng. and Ph.D. degrees in information and communication from Nanyang Technological University, Singapore, in 2000 and 2004, respectively. He was a Post-Doctoral Fellow with Lucent Technologies Bell Labs, Murray Hill, in 2005, and a Visiting Assistant Professor with The Hong Kong Polytechnic University in 2008. From 2006 to 2010, he was with the Institute for Infocomm Research, Singapore. From 2010 to 2023, he was with the Singapore University of Technology and Design. Since 2023, he has been with the School of Electrical and Electronic Engineering, Nanyang Technological University. He is a Distinguished Lecturer of IEEE Vehicular Technology Society, Top 2\% Scientists by Stanford University, and also a Highly Cited Researcher by Clarivate Web of Science. He currently serves as an Editor-in-Chief for Springer Nature Computer Science, Editor for the IEEE Transactions on Vehicular Technology, IEEE System Journal, and IEEE Transactions on Network Science and Engineering. He also served as the Guest Editor for several special issues, including IEEE Journal on Selected Areas in Communications, IEEE Wireless Communications Magazine, IEEE Transactions on Cognitive Communications and Networking, and Elsevier Applied Energy.
\end{IEEEbiography}

\vspace{-1.8cm}

\begin{IEEEbiography}[{\includegraphics[width=1in,height=1.25in,clip,keepaspectratio]{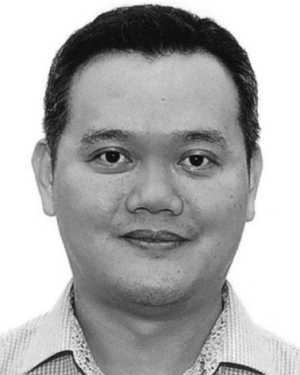}}]{U-Xuan Tan}
(Senior Member, IEEE) received the B.Eng. and Ph.D. degrees from Nanyang Technological University, Singapore, in 2005 and 2010, respectively. From 2009 to 2011, he was a Postdoctoral Fellow with the University of Maryland, College Park, MD, USA. From 2012 to 2014, he was a Lecturer with the Singapore University of Technology and Design, Singapore. He took up a research-intensive role as an Assistant Professor in 2014. He has been promoted to an Associate Professor in 2021. His research interests include mechatronics, sensing and control, onsite micromanipulation and microsensing, sensing and control technologies for human–robot interaction, and interdisciplinary teaching.
\end{IEEEbiography}


\begin{thebibliography}{99}

\bibitem{Introduction-multi-robot-system}
Z. Feng, G. Hu, Y. Sun, and J. Soon. "An overview of collaborative robotic manipulation in multi-robot systems," \textit{Annual Reviews in Control}, vol. 49, pp. 113-127, 2020.


\bibitem{Introduction-CADRE}
Cooperative Autonomous Distributed Robotic Exploration. [Online]. Available: https://www.nasa.gov/cooperative-autonomous-distributed-robotic-exploration-cadre/.

\bibitem{Introduction-decision-making}
K. Pongsirijinda, Z. Cao, M. Shalihan, B. K. K. Ng, B. P. L. Lau, C. Yuen, and U.-X. Tan, "GMC-Pos: Graph-Based Multi-Robot Coverage Positioning Method," \textit{2023 IEEE International Conference on Robotics and Biomimetics (ROBIO)}, Koh Samui, Thailand, 2023, pp. 1-7.

\bibitem{Introduction-cooperative-operation}
P.Y. Lajoie and G. Beltrame, "Swarm-SLAM: Sparse Decentralized Collaborative Simultaneous Localization and Mapping Framework for Multi-Robot Systems," \textit{IEEE Robotics and Automation Letters}, vol. 9, no. 1, pp. 475-482, Jan. 2024.

\bibitem{Introduction-relative-category}
A. Eskandarian, C. Wu, and C. Sun, "Research Advances and Challenges of Autonomous and Connected Ground Vehicles," \textit{IEEE Transactions on Intelligent Transportation Systems}, vol. 22, no. 2, pp. 683-711, Feb. 2021.


\bibitem{Introduction-centralized}
M. W. Mehrez, I. Mann, K. George, and R. G. Gosine, "An optimization based approach for relative localization and relative tracking control in multi-robot systems," \textit{Journal of Intelligent \& Robotic Systems}, vol. 85, pp. 385-408, 2017.


\bibitem{Introduction-distributed}
Y. Liu, Y. Wang, J. Wang, and Y. Shen, "Distributed 3D Relative Localization of UAVs," \textit{IEEE Transactions on Vehicular Technology}, vol. 69, no. 10, pp. 11756-11770, Oct. 2020.

\bibitem{Introduction-LiDAR-camera}
H. Song, W. Choi, and H. Kim, "Robust Vision-Based Relative-Localization Approach Using an RGB-Depth Camera and LiDAR Sensor Fusion," \textit{IEEE Transactions on Industrial Electronics}, vol. 63, no. 6, pp. 3725-3736, Jun. 2016.

\bibitem{Introduction-robot-id}
E. Héry, P. Xu, and P. Bonnifait, "Consistent decentralized cooperative localization for autonomous vehicles using LiDAR, GNSS, and HD maps," \textit{Journal of Field Robotics}, vol. 38, no. 4, pp. 552--571, 2021.



\bibitem{Introduction-UWB}
A. R. Jiménez Ruiz and F. Seco Granja, "Comparing Ubisense, BeSpoon, and DecaWave UWB Location Systems: Indoor Performance Analysis," \textit{IEEE Transactions on Instrumentation and Measurement}, vol. 66, no. 8, pp. 2106-2117, Aug. 2017.

\bibitem{Introduction-UWB-2}
M. Shalihan, R. Liu, and C. Yuen, "NLOS Ranging Mitigation with Neural Network Model for UWB Localization," \textit{2022 IEEE 18th International Conference on Automation Science and Engineering (CASE)}, Mexico City, Mexico, 2022, pp. 1370-1376.


\bibitem{AoA-1}
L. Taponecco, A. A. D'Amico, and U. Mengali, "Joint TOA and AOA Estimation for UWB Localization Applications," \textit{IEEE Transactions on Wireless Communications}, vol. 10, no. 7, pp. 2207-2217, Jul. 2011.

\bibitem{AoA-2}
S. Wang, X. Jiang, and H. Wymeersch, "Cooperative Localization in Wireless Sensor Networks With AOA Measurements," \textit{IEEE Transactions on Wireless Communications}, vol. 21, no. 8, pp. 6760-6773, Aug. 2022.



\bibitem{Introduction-adaptive-clustering}
Z. Yan, T. Duckett, and N. Bellotto, "Online learning for 3D LiDAR-based human detection: Experimental analysis of point cloud clustering and classification methods," \textit{Autonomous Robots}, vol. 44, pp. 147--164, 2020.

\bibitem{Introduction-matching}
C. McCreesh, P. Prosser, and J. Trimble, "A Partitioning Algorithm for Maximum Common Subgraph Problems," \textit{Proceedings of the 26th International Joint Conference on Artificial Intelligence (IJCAI'17)}, Melbourne, Australia, 2017, pp. 712–719.

\bibitem{Introduction-euclidean-similarity}
G. Qian, S. Sural, Y. Gu, and S. Pramanik, "Similarity between Euclidean and cosine angle distance for nearest neighbor queries," \textit{Proceedings of the 2004 ACM symposium on Applied computing}, New York, NY, USA, 2004, pp. 1232--1237.


\bibitem{Related-work-exploration}
B. P. L. Lau, B. J. Y. Ong, L. K. Y. Loh, R. Liu, C. Yuen, G. S. Soh, and U.-X. Tan, "Multi-AGV's Temporal Memory-Based RRT Exploration in Unknown Environment," \textit{IEEE Robotics and Automation Letters}, vol. 7, no. 4, pp. 9256-9263, Oct. 2022.

\bibitem{Related-work-formation}
J. Alonso-Mora, E. Montijano, M. Schwager, and D. Rus, "Distributed multi-robot formation control among obstacles: A geometric and optimization approach with consensus," \textit{2016 IEEE International Conference on Robotics and Automation (ICRA)}, Stockholm, Sweden, 2016, pp. 5356-5363.

\bibitem{Related-work-distributed-survey}
S. Wang, Y. Wang, D. Li, and Q. Zhao, "Distributed Relative Localization Algorithms for Multi-Robot Networks: A Survey," \textit{Sensors}, vol. 23, no. 5, pp. 2399, 2023.


\bibitem{Related-work-visual-01}
R. T. Rodrigues, P. Miraldo, D. V. Dimarogonas, and A. P. Aguiar, "A Framework for Depth Estimation and Relative Localization of Ground Robots using Computer Vision," \textit{2019 IEEE/RSJ International Conference on Intelligent Robots and Systems (IROS)}, Macau, China, 2019, pp. 3719-3724.


\bibitem{Related-work-visual-02}
M. Karrer, M. Agarwal, M. Kamel, R. Siegwart, and M. Chli, "Collaborative 6DoF Relative Pose Estimation for Two UAVs with Overlapping Fields of View," \textit{2018 IEEE International Conference on Robotics and Automation (ICRA)}, Brisbane, QLD, Australia, 2018, pp. 6688-6693.



\bibitem{Related-work-visual-2}
M. Pavliv, F. Schiano, C. Reardon, D. Floreano, and G. Loianno, "Tracking and Relative Localization of Drone Swarms With a Vision-Based Headset," \textit{IEEE Robotics and Automation Letters}, vol. 6, no. 2, pp. 1455-1462, Apr. 2021.


\bibitem{Related-work-yolov3}
J. Redmon and A. Farhadi, “Yolov3: An incremental improvement,” 2018,
https://arxiv.org/abs/1804.02767.




\bibitem{Related-work-visual-3}
Z. Xun, J. Huang, Z. Li, Z. Ying, Y. Wang, C. Xu, F. Gao, and Y. Cao, "CREPES: Cooperative Relative Pose Estimation System." \textit{2023 IEEE/RSJ International Conference on Intelligent Robots and Systems (IROS)}, Detroit, MI, USA, 2023, pp. 5274-5281.


\bibitem{Related-work-DS}
V. Usenko, N. Demmel, and D. Cremers, "The Double Sphere Camera Model," \textit{2018 International Conference on 3D Vision (3DV)}, Verona, Italy, 2018, pp. 552-560.


\bibitem{Related-work-lidar-clustering}
F. Gao, C. Li, and B. Zhang, "A Dynamic Clustering Algorithm for Lidar Obstacle Detection of Autonomous Driving System," \textit{IEEE Sensors Journal}, vol. 21, no. 22, pp. 25922-25930, Nov. 2021.


\bibitem{Related-work-lidar-1}
F. Zhu, Y. Ren, F. Kong, H. Wu, S. Liang, N. Chen, W. Xu, and F. Zhang, "Swarm-LIO: Decentralized Swarm LiDAR-inertial Odometry," \textit{2023 IEEE International Conference on Robotics and Automation (ICRA)}, London, United Kingdom, 2023, pp. 3254-3260.

\bibitem{Related-work-lidar-2}
M. Shalihan, Z. Cao, K. Pongsirijinda, L. Guo, B. P. L. Lau, R. Liu, C. Yuen, and U.-X. Tan, "Moving Object Localization based on the Fusion of Ultra-WideBand and LiDAR with a Mobile Robot," \textit{2023 IEEE International Conference on Robotics and Biomimetics (ROBIO)}, Koh Samui, Thailand, 2023, pp. 1-8,


\bibitem{Related-work-radio}
H. Wymeersch, J. He, B. Denis, A. Clemente, and M. Juntti, "Radio Localization and Mapping With Reconfigurable Intelligent Surfaces: Challenges, Opportunities, and Research Directions," \textit{IEEE Vehicular Technology Magazine}, vol. 15, no. 4, pp. 52-61, Dec. 2020.

\bibitem{Related-work-uwb-1}
G. Schroeer, “A real-time UWB multi-channel indoor positioning system for industrial scenarios,” \textit{2018 International Conference on Indoor Positioning and Indoor Navigation (IPIN)}, Nantes, France, 2018, pp. 1-5.

\bibitem{Related-work-uwb-2}
Z. Cao, R. Liu, C. Yuen, A. Athukorala, B.K.Kiat Ng, M. Mathanraj, and U.-X. Tan "Relative Localization of Mobile Robots with Multiple Ultra-WideBand Ranging Measurements," \textit{2021 IEEE/RSJ International Conference on Intelligent Robots and Systems (IROS)}, Prague, Czech Republic, 2021, pp. 5857-5863.

\bibitem{Related-work-uwb-3}
R. Liu, Z. Deng, Z. Cao, M. Shalihan, B. P. L. Lau, K. Chen, K. Bhowmik, C. Yuen, and U.-X. Tan, "Distributed Ranging SLAM for Multiple Robots with Ultra-WideBand and Odometry Measurements," \textit{2022 IEEE/RSJ International Conference on Intelligent Robots and Systems (IROS)}, Kyoto, Japan, 2022, pp. 13684-13691.


\bibitem{Related-work-uwb-4}
T. H. Nguyen and L. Xie, "Relative Transformation Estimation Based on Fusion of Odometry and UWB Ranging Data," \textit{IEEE Transactions on Robotics}, vol. 39, no. 4, pp. 2861-2877, Aug. 2023.



\bibitem{graph-theory}
D. B. West, "Introduction to graph theory," Upper Saddle River: Prentice hall, vol. 2, 2001.



\bibitem{system-max-likelihood}
S. Thrun, W. Burgard, and D. Fox, "Probabilistic Robotics," MIT Press, 2005.


\bibitem{experiment-ros}
M. Quigley, K. Conley, B. Gerkey, J. Faust, T. Foote, J. Leibs, R. Wheeler, and A. Y. Ng, "ROS: an open-source Robot Operating System," \textit{ICRA workshop on open source software}, vol. 3, no. 3.2, pp. 5, 2009.


\bibitem{UWB-stability}
J. Cano, G. Pagès, É. Chaumette, and J. LeNy, "Clock and Power-Induced Bias Correction for UWB Time-of-Flight Measurements," \textit{IEEE Robotics and Automation Letters}, vol. 7, no. 2, pp. 2431-2438, April 2022.


\end{thebibliography}
\end{document}